%% file: sample-manuscript.tex
\documentclass[manuscript, screen]{acmart}

\usepackage{hyperref}

\usepackage{url}
\usepackage[T1]{fontenc}
\usepackage{graphicx}
\usepackage{subcaption}
\usepackage{multirow}
\usepackage{color}
\usepackage{booktabs}
\usepackage{tabularray}
\usepackage{float}
\usepackage{tcolorbox}
\usepackage{soul}
\usepackage[bottom]{footmisc}
\usepackage{wrapfig}
\usepackage{lipsum}
\usepackage{colortbl}

\AtBeginDocument{%
  }

\acmJournal{TALLIP}
\begin{document}

\title{From Human Judgements to Predictive Models: Unravelling Acceptability in Code-Mixed Sentences}


\author{Prashant Kodali}
\email{prashant.kodali@research.iiit.ac.in}
\affiliation{%
  \institution{IIIT Hyderabad}
  \city{Hyderabad}
  \state{Telengana}
  \country{India}
}

\author{Anmol Goel}
\affiliation{%
  \institution{IIIT Hyderabad}
  \city{}
  \country{India}}
\additionalaffiliation{%
  \institution{UKP Lab, Technical University of Darmstadt}
  \city{Darmstadt}
  \country{Germany}
}  
  
\email{anmol.goel@research.iiit.ac.in}

\author{Likhith Asapu}
\affiliation{%
  \institution{IIIT Hyderabad}
  \city{}
  \country{India}}
\email{likhith.a@research.iiit.ac.in}

\author{Vamshi Krishna Bonagiri}
\affiliation{%
  \institution{IIIT Hyderabad}
  \city{}
  \country{India}}
\email{vamshi.b@research.iiit.ac.in}

\author{Anirudh Govil}
\affiliation{%
  \institution{IIIT Hyderabad}
  \city{}
  \country{India}}
\email{anirudh.govil@students.iiit.ac.in}

\author{Monojit Choudhury}
\affiliation{%
  \institution{MBZUAI}
  \city{Abu Dhabi}
  \country{United Arab Emirates}}
\email{monojitc@microsoft.com}

\author{Ponnurangam Kumaraguru}
\affiliation{%
  \institution{IIIT Hyderabad}
  \city{}
  \country{India}}
\email{pk.guru@iiit.ac.in}

\author{Manish Shrivastava}
\affiliation{%
  \institution{IIIT Hyderabad}
  \city{}
  \country{India}}
\email{m.shrivastava@iiit.ac.in}

\authorsaddresses{%
Authors' addresses: Prashant Kodali, Anmol Goel, Likhith Asapu, Vamshi Krishna Bonagiri, Anirudh Govil, Manish Shrivastava, Ponnurangam Kumaraguru, Address: IIIT Hyderabad, Prof CR Rao Road, Hyderabad, Telengana, India, 500032, E-Mail: \{prashant.kodali, anmol.goel, likhith.a, vamshi.b\}@research.iiit.ac.in, anirudh.govil@students.iiit.ac.in, \{m.shrivastava, pk.guru\}@iiit.ac.in ; Monojit Choudhury, Address: MBZUAI, Masdar City, Abu Dhabi, United Arab Emirates, Emails: monojit.choudhury@gmail.com}

\renewcommand{\shortauthors}{Kodali, et al.}

\makeatletter
\renewcommand{\thesubfigure}{(\alph{subfigure})}
\makeatother

\begin{abstract}
  \input{sections/00-abstract.tex}
\end{abstract}






\received{2 December 2023}
\received[revised]{18 October 2024}
\received[revised]{25 February 2025}
\received[accepted]{5 May 2025}

\maketitle

\input{sections/01-intro.tex}

\input{sections/02-background.tex}

\input{sections/03-data.tex}
\input{sections/04-dataset-Analysis.tex}

\input{sections/05-classifier.tex}
\input{sections/06-conclusion.tex}

\begin{acks}
We gratefully acknowledge the support of Microsoft Research India through the MSR India PhD Award grant. We extend our sincere thanks to Karuna Chandra, Bhaskara Hanuma V, Ritwik Mishra, Aparajitha A., and Ananya Mukherjee for their thoughtful reviews and constructive feedback. We also thank our fellow labmates for their insightful discussions and continuous support throughout the course of this work.
\end{acks}

\bibliographystyle{ACM-Reference-Format}
\bibliography{bibliography}


\appendix
\input{sections/appendix}

\end{document}

%% file: sections/00-abstract.tex
Current computational approaches for analysing or generating code-mixed sentences do not explicitly model ``naturalness'' or ``acceptability'' of code-mixed sentences, but rely on training corpora to reflect distribution of acceptable code-mixed sentences. Modelling human judgement for the acceptability of code-mixed text can help in distinguishing natural code-mixed text and enable quality-controlled generation of code-mixed text. To this end, we construct Cline - a dataset containing human acceptability judgements for English-Hindi~(en-hi) code-mixed text. Cline is the largest of its kind with 16,642 sentences, consisting of samples sourced from two sources: synthetically generated code-mixed text and samples collected from online social media. Our analysis establishes that popular code-mixing metrics such as CMI, Number of Switch Points, Burstines, which are used to filter/curate/compare code-mixed corpora have low correlation with human acceptability judgements, underlining the necessity of our dataset. Experiments using Cline demonstrate that simple Multilayer Perceptron (MLP) models when trained solely using code-mixing metrics as features are outperformed by fine-tuned pre-trained Multilingual Large Language Models (MLLMs). Specifically, among Encoder models XLM-Roberta and Bernice outperform IndicBERT across different configurations. Among Encoder-Decoder models, mBART performs better than mT5, however Encoder-Decoder models are not able to outperform Encoder-only models. Decoder-only models perform the best when compared to all other MLLMS, with Llama 3.2 - 3B models outperforming similarly sized Qwen, Phi models. Comparison with zero and fewshot capabilitites of ChatGPT show that MLLMs fine-tuned on larger data outperform ChatGPT, providing scope for improvement in code-mixed tasks. Zero-shot transfer from English-Hindi to English-Telugu acceptability judgments using our model checkpoints proves superior to random baselines, enabling application to other code-mixed language pairs and providing further avenues of research. We publicly release our human-annotated dataset, trained checkpoints, code-mix corpus, and code for data generation and model training.

%% file: sections/01-intro.tex
\section{Introduction}\label{sec:01_intro}

Code-Switching/Code-Mixing\footnote{Code-Switching usually refers to inter-sentence mixing, and Code-Mixing is used to convey intra-sentence mixing. In this work, we use Code-Mixing as an umbrella term for all language mixing phenomena.} is the phenomenon of mixing linguistic units from two or more languages, and multilingual users do so because of various psycho- and socio-linguistic reasons~\citep{gumperz1977sociolinguistic}. Linguistic research has shown that code-mixing is not an arbitrary mixing of tokens from different languages, and multi-lingual users possess strong judgement vis-a-vis an ``acceptable'' vs ``unacceptable'' code-mixed sentence~\citep{joshi-1982-processing}. Linguists have also proposed certain syntactic constraints that govern how languages mix to create ``valid'' code-mixed sentences, Equivalence Constraint~(EC) Theory~\citep{poplack-1980-sometimes}, Matrix Language~(ML) Theory~\citep{myers1997duelling}, to name a few. Such theories have also been utilized by computational tools to generate code-mixed text. However, the universality of such grammatical constraint is limited, as evidenced by counter-examples where grammatical theories based on constraints do not hold.~\citep{Muysken2000BilingualSA,Sciullo1986GovernmentAC}. 

Recent advancements in data-centric NLP methods have emphasised the importance of large corpora for pre-training, demonstrating their potential to yield favourable results in downstream tasks. To extend the same to code-mixed settings, limited availability of code-mixed corpora, both in quantity and quality, is an impediment. Code-mixed text is not prevalent in traditional sources of text corpora (e.g. Wikipedia, News), but quite frequent in informal speech events, inter-personal messaging applications~(e.g. WhatsApp) and social networks. Collecting code-mixed data at scale from such sources is not trivial and is usually accompanied by noise and/or source-specific quirks~(URLs, Hashtags, user mentions)~\cite{cetinoglu-etal-2016-challenges}. Due to the limited availability of code-mixed data~\cite{sitaram2020survey}, computational studies on code-mixed text have relied on generating synthetic code-mixed text to augment data~\citep{Gupta2021TrainingDA,Winata2019CodeSwitchedLM}. Synthetic code-mixed text corpora can be leveraged for data augmentation only if quality of augmented text is ensured. Ensuring that the generated code-mixed text is considered ``acceptable'' by native human speakers is crucial for maintaining data quality. Moreover, it is essential for models generating code-mixed sentences to produce only naturally occurring code-mixed sentences, as this greatly enhances their usefulness.

We use the following example to motivate our study.\footnote{Bold tokens are Hindi words written in roman script, gloss is provided in brackets.}
\begin{enumerate}\label{motivating_example}
    \item[(1)] I conduct research on code-mixing. \label{motivating_example_sent1}
    \item[(2)] \textbf{main}~\textit{(I)} code-mixing \textbf{par}~\textit{(on)} research \textbf{karta hoon}~\textit{(do)}. \label{motivating_example_sent2}
    \item[(3)] I do \textbf{shodhkarya}~\textit{(research)} on code-mixing.\label{motivating_example_sent3}
    \item[(4)] * \textbf{main}~\textit{(I)} conduct research code-mixing \textbf{par}~\textit{(I)}. \label{motivating_example_sent4}
\end{enumerate}

Sentence (1) is in monolingual English, and examples (2), (3), and (4) represent different ways of rewriting (1) as a code-mixed sentence. We note that examples (2) to (4) exhibit varying degrees of acceptability for a code-mixed sentence—(2) is deemed acceptable, (4) is clearly unacceptable, and (3) is syntactically correct but unnatural. Given that multilingual speakers possess strong judgments about how to mix two languages~\cite{joshi-1982-processing}, understanding and modelling acceptability of code-mixed text can catalyse quality-controlled data augmentation for improving natural code-mixed text generation, and downstream tasks. However, the resource-poor nature of code-mixed text poses a challenge in obtaining a general distribution of such ``acceptable'' sentences from large corpora.


 
Acceptability judgements have traditionally been employed as the principal means of data acquisition for constructing linguistic theories to understand human competency and performance in natural language. 
Recently, NLP community has focused on computational modelling of acceptability judgements and examining language models' capacity to differentiate between acceptable and unacceptable sentences. Computational evaluation of acceptability in monolingual settings has been tackled through corpora such as CoLA~\citep{Warstadt2018NeuralNA} and BLiMP~\citep{warstadt-etal-2020-blimp-benchmark} for English, ItaCoLA~\citep{trotta-etal-2021-monolingual-cross} for Italian, and CLiMP~\citep{xiang-etal-2021-climp} for Chinese. However, for code-mixed text, acceptability and its computational modelling remain under-explored.

Traditionally, metrics like Code Mixing Index, Burstiness, Number of Switch Points~\cite{guzman17_interspeech, gamback-das-2016-comparing} have been used for assessing degree of code-mixing in a sentence/corpora, and for curating code-mixed data resources. While metrics of code-mixing can quantify ``degree'' of code-mixing, we hypothesise that such metrics have limited potential to distinguish between an acceptable and unacceptable sentence. The absence of a data resource reflecting human judgments on the acceptability of code-mixed text impedes the rigorous evaluation of existing code-mixing metrics and the computational modeling of acceptability, thereby creating a significant gap in assessing the acceptability of code-mixed text.

We situate our study within this gap in the literature. We collect human judgements on acceptability of code-mixed sentences to analyse the phenomena, and explore the possibility of predicting acceptability of code-mixed text. Following are the main contributions of our work:
\begin{itemize}
    \item We create a novel dataset, \textbf{Cline}, consisting of acceptability annotations for code-mixed English-Hindi~(en-hi) sentences sourced from online social networks (user-generated text) and synthetic code-mix generation tools.
    \item Our analysis of Cline empirically provides support to the argument that acceptability of code-mixed text lies in a gradient of human acceptability judgements, and bilinguals possess strong judgements regarding which code-mixed sentences are acceptable. Our analysis shows that conventional code-mixing metrics such as Code Mixing Index, Switch Points, Burstiness demonstrate limited correlation with human judgements, further highlighting the significance of Cline. This finding has implications for how code-mixed datasets are curated and processed.
    \item We also provide strong baselines for predicting the acceptability judgements of code-mixed text. Our results show that fine-tuned Multilingual Large Language Models outperform predictive models trained using code-mixing metrics as features and human baselines. We compare our model's performance with instruction following interactive model, ChatGPT, and our results indicate that our fine-tuned models fare better in comparison. Our results for zero-shot transfer of acceptability prediction to English-Telugu~(en-te) data have implications for the transfer of acceptability judgements to different code-mixed language pairs in challenging settings and provide preliminary evidence of performance gains on unseen target languages. 
\end{itemize}


The remainder of the paper is structured as follows: In the following section, we discuss
prior research on computational approaches to acceptability and code-mixing acceptability.
In Section~\ref{sec:03_data}, we present our data curation and annotation methodology. In Section~\ref{sec:04_dataset_analysis}, we present a detailed analysis of our dataset, and insights from our systematic study of relationship between human acceptability judgement and existing code-mix metrics that can be used to judge acceptability of code-mixed sentences. In Section~\ref{sec:05_classifier}, we present set of experiments for predicting acceptability of code-mixed sentences. Finally, we provide a summary of our results, limitation of our study and possible avenues of future work in Section~\ref{sec:06_conclusion}.

Complete datasets along with the trained checkpoints can be found here: \protect\url{https://huggingface.co/collections/prakod/acceptability-of-code-mixed-text-67768246afb60d23f4003435}. Our codebase for creating the datasets, analysing the datasets, training models can be found here: \protect\url{https://github.com/prashantkodali/CodeMixAcceptability}.

%% file: sections/02-background.tex
\section{Acceptability and Code-mixing}\label{sec:02_background}




Code-Mixing of languages in natural conversations is an intrinsic part of a multilingual speaker's linguistic competency. Generating ``acceptable'' code-mixed sentences and enabling NLP pipelines to analyse code-mixed has long been of interest to NLP community. Here, we discuss two aspects central to this study - code-mixing and acceptability - and the relationship between the two.

\textbf{Quantifying Code-Mixing} In code-mixed text, linguistic units from multiple languages are interwoven, with speakers employing various mixing methods such as borrowing words and syntactical constructions. Consequently, code-mixing patterns can differ among speakers, sources, or corpora. Quantifying these patterns is essential for identifying, grading, and comparing code-mixed sentences.~\citet{guzman17_interspeech} provide a comprehensive summary of various code-mixing metrics. The metrics presented in this study rely on the language ID of the tokens in a code-mixed sentence, looking at the distribution of the language-wise tokens from multiple standpoints - ratio, temporal distribution of the two languages in a sentence. ~\citet{kodali-etal-2022-symcom} propose metric SyMCoM, which takes the language ID the PoS tag of the tokens in a sentence, providing means to quantify syntactic nature of switching. We refer readers to Appendix~\ref{appendix:codemix_metrics} for further details on the code-mixing metrics.

\input{tables/08_examples_metrics}
 Code-mixing metrics were designed to determine the ``degree'' of code-mixing, to curate datasets for code-mixed tasks. Code-mixing metrics usually analyse the distribution of token wise language IDs, and give quantitative means to grade code-mixed text. This design suggests that code-mixing metrics may not be fully adequate for evaluating the acceptability
of code-mixed sentences. However, such code-mixed metrics remain the only means of grading code-mixed corpora currently, thus necessitating a critical evaluation of their effectiveness in rating the acceptability of code-mixed utterances. Table~\ref{tab:examples_codemixmetrics} shows values of aforementioned code-mixing metrics for demonstrative example sentences (2)-(4) in Section~\ref{sec:01_intro}. These sentences differ have different patterns of mixing languages - a) alternating multiple times within a sentence; b) switching nouns as opposed to verbs/auxiliaries; c) switching all or some of the words with certain PoS tag. Code-mixing metrics are able to capture different ``signatures'' of code-mixing - for instance, SyMCoM scores change when different PoS categories are switched, CMI and Number of switch points change depending on how many tokens are switched, and Burstiness changes based on how the switch points are spread across a sentence. However, the challenge lies in determining if code-mixing metrics are sufficient to rate acceptability of code-mixed sentences.

\textbf{Acceptability} Computational modelling of acceptability judgements and probing language model's propensity towards acceptable sentences has gained the attention of the NLP community over past few years. Researchers have followed various paradigms to collect data for acceptability judgements: (a) minimal pairs; (b) binary judgements; (c) ratings on the Likert scale. In the case of minimal pairs, pairs of sentences are generated, which vary only on a particular rule (e.g. Subj-Verb agreement)~\citep{warstadt-etal-2020-blimp-benchmark}. For binary scale, expert judgements on examples are collected from previously published linguistics literature, and a corpus of acceptability is assembled~\citep{Warstadt2018NeuralNA}. 

While many sentences are either clearly acceptable or clearly unacceptable, a significant number of sentences fall somewhere in between in a grey area of partial acceptability~\citep{Lau2017GrammaticalityAA, Sprouse2007ContinuousAC}. Crowd-sourcing has emerged as a valid way of collecting  grammatical acceptability ratings, with recent evidence highlighting the reliability of crowd-sourced ratings with respect to tests conducted under laboratory conditions~\citep{Sprouse2007ContinuousAC}. \citet{lau2014measuring} employed variations in the mode of presentation - binary scale, four category scale, sliding scale - and showed that the aggregated rating is robust and is not affected by mode of presentation.

\textbf{Acceptability Judgements for code-mixed sentences}\label{02_sec:bacground-acceptability-codemix} is also dependent on multiple factors - exposure to code-mixing, speaker's level of proficiency in individual languages, and demographics (age, social status)~\citep{lederberg1985code}. There could also be multiple ways of writing two sentences~(demonstrated in example sentences (1)-(4) in Section~\ref{sec:01_intro}). Thus, the question of acceptability and handling it computationally for code-mixed sentences becomes a more challenging proposition. Linguistic studies have sought human judgments for code-mixed sentences~\citep{lederberg1985code}, wherein specific code-mixed constructions are studied on a smaller scale - small in terms of the number of sentences as well as the number of raters. 
 
 

Recently, \citet{srivastava-singh-2021-quality} conducted a shared task for predicting the quality of English-Hindi code-mixed sentences, where sentences were rated on a scale of 1-10. A limitation of this data is methodology used to create samples for human annotations: using word/phrase alignment of a parallel sentence to replace word/phrase in a sentence, which has limited coverage of code-mixing patterns. \citet{pratapa-choudhury-2021-comparing} conducted a human evaluation of two popular grammatical theories, Matrix Language~(ML) and Equivalence Constraint (EC), to assess which grammatical theory is preferred by English-Spanish bilinguals. This study, however, did not focus on analysis or computational modelling of acceptability for code-mixed text, instead, explored human preference among the grammatical theories of code-mixing.~
The authors observed no clear winner for acceptability between the two grammatical theories, thus highlighting the need for a more nuanced analysis of which code-mixed sentences are acceptable and leverage a gradient-based approach to capture human judgements on acceptability of code-mixed sentences instead of a binary preference scale. Both studies are limited by the number of annotated samples (1,976 samples in \citet{srivastava-singh-2021-quality}, 429 samples in \citet{pratapa-choudhury-2021-comparing}) and the syntactic/patterns of switching in the samples. Motivated by the limitations of prior work on code-mixed datasets, we undertake the current study to create a larger dataset covering diverse patterns of code-mixing, and capture human judgements on their acceptability to enable efficient modelling of code-mixed acceptability judgements.




%% file: tables/08_examples_metrics.tex
\begin{table}[h]
\caption{A comparative evaluation of code-mixing metrics against sentences with varying degree of acceptability. Among the example sentences, (2) is more acceptable than (3) and (3) is more acceptable than (4). However, Code-mixing metrics exhibit no discernible pattern with respect to acceptability of these sentences.}
\label{tab:examples_codemixmetrics}
\centering
\resizebox{\linewidth}{!}{%
\begin{tblr}{
  cells = {c},
  cell{1}{1} = {r=2}{},
  cell{1}{2} = {r=2}{},
  cell{1}{3} = {r=2}{},
  cell{1}{4} = {r=2}{},
  cell{1}{5} = {c=5}{},
  hline{1,3,6} = {-}{},
  hline{2} = {5-9}{},
}
\textbf{Sentence }                                                                                                    & \textbf{CMI } & {\textbf{Num of}\\\textbf{Switch Points}} & \textbf{Burstiness} & \textbf{SyMCoM} &               &               &              &               \\
                                                                                                                      &               &                                           &                     & \textbf{NOUN}           & \textbf{PREP} & \textbf{VERB} & \textbf{AUX} & \textbf{Sent} \\
{(2) \textbf{main}~\textit{(I)} code-mixing \textbf{par}~\textit{(on)}~\\research \textbf{karta hoon}~\textit{(do)}.} & 33.3          & 4                                         & -0.45               & -0.33                   & 1             & 1             & 1            & 0.66          \\
{(3) I do \textbf{shodhkarya}~\textit{(research)}~\\on code-mixing.}                                                  & 20            & 2                                         & -0.48               & -0.33                   & -1            & -1            & NA           & 0.59          \\
{(4) \textbf{main}~\textit{(I)} conduct research~\\code-mixing \textbf{par}~\textit{(I)}.}                            & 40            & 2                                         & -0.17               & -0.33                   & 1             & -1            & NA           & 0.59          
\end{tblr}
}

\end{table}

%% file: sections/03-data.tex
\section{Cline - A Dataset for Acceptability of Code-Mixing}\label{sec:03_data}

\begin{figure}[!t]
    \centering
    \includegraphics[width=0.9\linewidth]{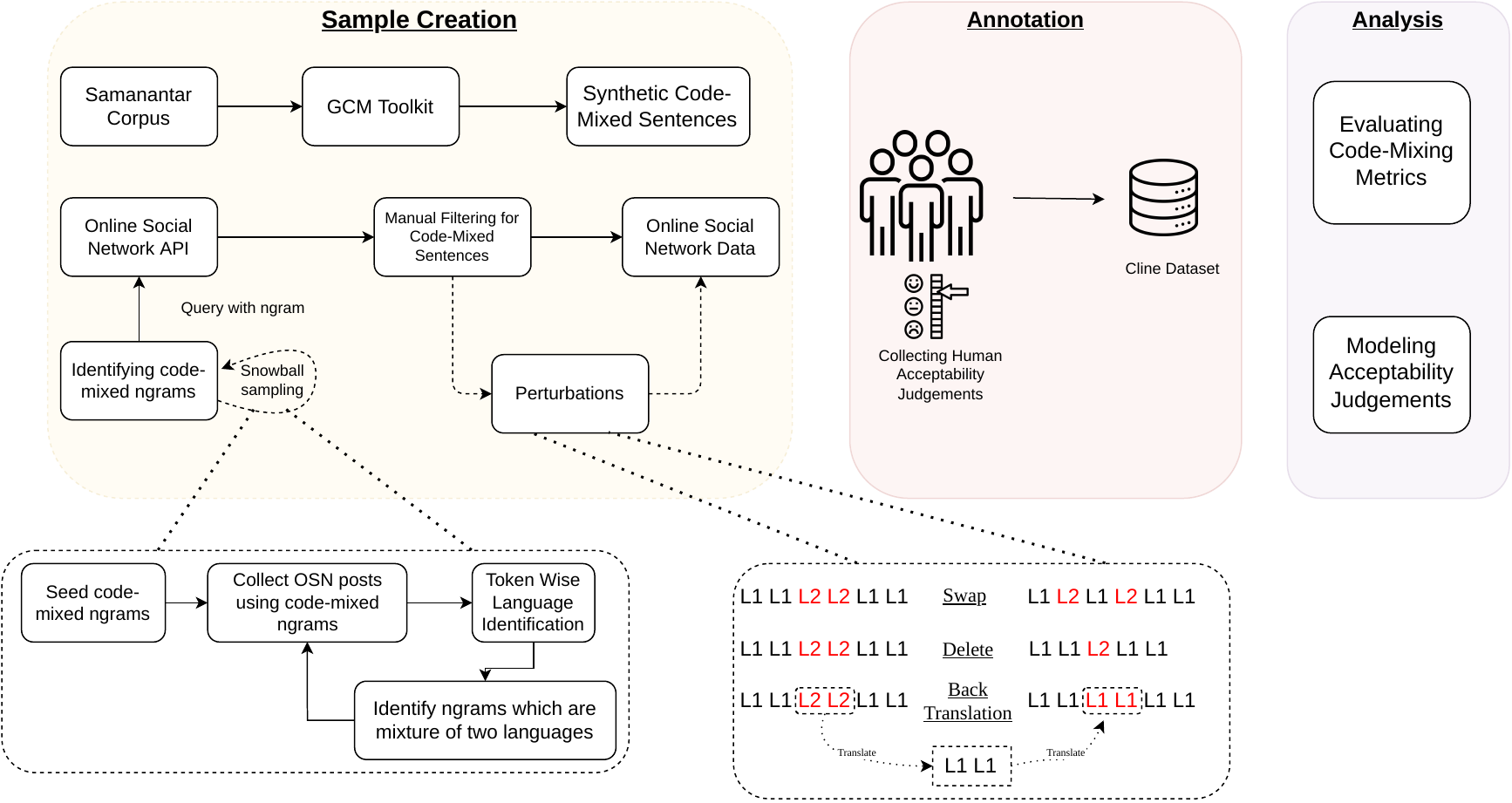}
    \caption{Overview of our study involving three prime components: a) creating code-mixed samples; b) collecting human judgments on acceptability; c) evaluating of code-mix metrics and training predictive models. For perturbations to OSN text L1, L2 indicate the language ID of the tokens in a code-mixed sentence~(e.g L1 could be Hindi, and L2 could be English).}
    \label{fig:overview}
\end{figure}

Prior work on acceptability have collated expert linguist's judgments from existing literature~\citep{Warstadt2018NeuralNA} or sampling sentences from existing monolingual corpora~\citep{Lau2017GrammaticalityAA} for curating samples for monolingual acceptability judgments datasets. Both of the aforementioned approaches are not feasible in the context of code-mixed English-Hindi - as expert judgments on code-mixed sentences are very limited, and large code-mixed corpora aren't publicly available from which sentences can be sampled. To overcome this we take a multi-pronged approach to carefully collect samples for human annotations - a) collect both human generated and synthetically generated code-mixed text , b) collect samples which cover a spectrum of various degrees of code-mixing. Figure\hbox{~\ref{fig:overview}} presents an overview of the study's scope, accompanied by a schematic illustrating the data collection and curation process.

\subsection{Curating samples for annotation} \label{sec:03_data_sample-creation}

To ensure the diversity in the patterns of code-mixing in our dataset, we collect sentences from two sources: a) User-generated code-mixed sentences from Online Social Network site - henceforth referred to as ``OSN''; b) Synthetic code-mixed sentences using GCM-Toolkit~\citep{rizvi-etal-2021-gcm} - henceforth referred to as ``GCM''.

\subsubsection{Social Network Text}\label{sec:03_data_OSN}
Traditionally, code-mixed datasets~\citep{khanuja-etal-2020-gluecos, aguilar-etal-2020-lince} have relied on sentences sourced from Online social networks, predominantly Twitter, Facebook, and YouTube comments. For our purposes, we focus on Twitter as our source of code-mixed text. We queried Twitter's API with specific query terms to collect code-mixed posts. To ensure that we are indeed collecting code-mixed sentences, we used code-mixed n-grams~(e.g. ``i wish \textbf{ki}~(\textit{that})'') as our query terms. To ensure diversity in the collected samples, the list of code-mixed query n-grams was expanded using snowball sampling. To start with we manually constructed 5 seed query code-mixed 3-gram (e.g. ``\textbf{hai}~(\textit{is}) but I''), using which we collected tweets. On the collected tweets, we computed token wise Language ID~(LID) using pre-trained tool released by \cite{bhat-etal-2018-universal}. Based on LIDs, we recomputed code-mixed bi-grams, tri-grams, 4-grams. Using the computed n-grams we again queried Twitter API. We did four such iterations, ending up with large corpus of tweets. For API query using a particular query term, we limited number of tweets, to ensure that one query term doesn't dominate the corpus. We randomly selected a set of samples from the collected corpus, and further manually filtered out samples that were abusive, monolingual, or contained language other than en or hi, thus ensuring that our dataset has proper code-mixed sentences. OSN samples have quirks and noise of social networks (user mentions, URLs, hashtags, emojis, non-standard punctuation, capitalisation, spelling, vocabulary, and syntax~\cite{eisenstein-2013-bad}).


\textit{Perturbing Text from online social network}: Since the collected social network posts are human-generated, which are likely to be on the acceptable end of the acceptability spectrum, to ensure the dataset contains samples across the spectrum of acceptability we artificially create negative samples by introducing perturbations to a subset of code-mixed tweets. We perturb OSN samples our study to generate negative samples for the task of acceptability classification. Use of perturbations is well-established in the literature as effective approaches for creating negative samples in similar contexts\hbox{~\cite{Lau2014MeasuringGI, Mikhailov2022RuCoLARC, warstadt-etal-2019-neural}}, such as grammaticality judgment tasks or code-mixed text evaluation. Without employing such techniques, it would be challenging to construct a robust dataset that includes a spectrum of acceptable and unacceptable samples, as purely authentic unacceptable sentences would be sparse and skewed toward one end of the spectrum. We do not place the additional constraint that the negative samples need to be semantically valid, as such samples do not occur naturally in any text sources. Adding a requirement for semantic correctness to the perturbed sentences would unnecessarily complicate the sample generation process and shift the focus of the study away from acceptability judgments, which is the central investigation in our work. Additionally, we would like to emphasize that the hard negative samples in our dataset come from GCM framework~(Section \hbox{\ref{sec:03_data_GCM}}), which involves synthetically generating code-mixed sentences. These hard negatives represent structurally plausible but unacceptable sentences, ensuring that the model is exposed to nuanced distinctions.

We used three perturbation techniques: a) swapping tokens, b) deleting tokens, c) translating spans of a language span. The deletion perturbation technique involves removing tokens from the source sentence. For the swapping perturbation, random pairs of tokens were chosen from the source text and their positions were swapped. For Back Translation perturbation, we chose a span of words (belonging to a single language), translated it into another language and translated it back to the source language. This span was chosen such that it is the longest span of words belonging to a single language. We used the NLPAug library~\citep{ma2019nlpaug} to perform the deletion, and swap operations. We used Google Translate API to perform back translation perturbation on the data.


\subsubsection{Synthetically Generated Text}\label{sec:03_data_GCM}

\begin{figure}[!b]
    \centering
    \includegraphics[width=1\linewidth]{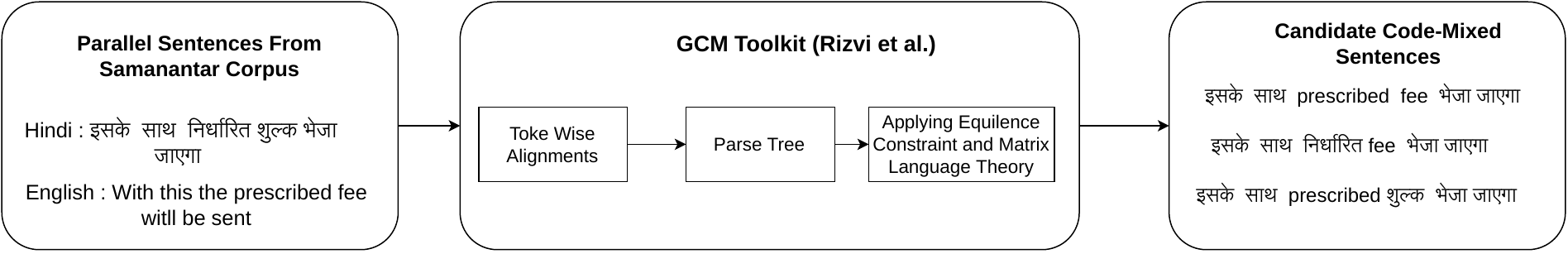}
    \caption{Illustrative example of how Samanantar corpus and GCM toolkit are used to create synthetic code-mixed sentences which are annotated by humans. We refer readers to \hbox{\cite{rizvi-etal-2021-gcm, Pratapa2018LanguageMF}} for a more detailed description of EC and ML theories, and their computational implementation for generating synthetic code-mixed sentences.}
    \label{fig:gcm-illustrative-example}
\end{figure}

Researchers have proposed various methods for synthetically generating code-mixed sentences - some of which are grounded in linguistic theories of code-mixing~\citep{Pratapa2018LanguageMF, Lee2019LinguisticallyMP}. Code-mixed generators that are based on grammatical theories are capable of generating a wider variety in code-mixed sentences - natural or not.  For the purpose of this study, we use GCM toolkit~\citep{rizvi-etal-2021-gcm} for producing synthetic code-mixed sentences. Figure\hbox{~\ref{fig:gcm-illustrative-example}} provides overview of the process leveraging the Samanantar parallel corpus and GCM toolkit for generating synthetic code-mixed sentences. We refer readers to \hbox{\cite{rizvi-etal-2021-gcm, Pratapa2018LanguageMF}} for a more detailed description of EC and ML theories, and their computational implementation for generating synthetic code-mixed sentences, as their detailed description is out of the scope of this work. GCM toolkit generates sentences based on Equivalence Constraint (EC)~\citep{poplack-1980-sometimes} and Matrix Language (ML)~\citep{myers1997duelling} theory. Input for GCM toolkit is a parallel corpus of the constituent language pair - in our case, a English - Hindi parallel corpus. We use Samanantar~\citep{ramesh-etal-2022-samanantar} as our en-hi monolingual parallel corpus and apply heuristics~\citep{mujadia-sharma-2022-ltrc} to filter out high-quality parallel sentences from Samanantar Dataset. For each input parallel sentence, GCM can generate multiple candidates, and we randomly sample from the generated sentences for our annotation. \newline

As samples in code-mixing benchmarks~\cite{khanuja-etal-2020-gluecos,aguilar-etal-2020-lince} are sourced from online social networks, the datasets are skewed towards a particular manner of writing code-mixed text. Dominance of social network samples in benchmarks can lead to a skewed assessment of model performance - model's subpar performance in code-mixed settings can be due to text being code-mixed or due to the social media style of writing that is substantially different from standard text. Sourcing data from two sources ensures diversity in the dataset. We recommend that a combination of both, synthetically generated sentences and human-generated social media text, should be utilised as a standard practice for curating samples in code-mixed settings.


\subsection{Crowd-sourcing Annotations- Annotation Guidelines and Setup}\label{sec:03_data_crowd_sourcing}

We base our annotation setup on the previous acceptability literature, which has established that acceptability judgments are gradient in nature i.e many sentences can be clearly acceptable/unacceptable, but a significant number of sentences fall somewhere in between in a grey area of partial acceptability~\citep{Lau2017GrammaticalityAA}. With the increased crowd-sourcing of human acceptability judgments, researchers have incorporated sophisticated non-binary scales like the Likert scale~\cite{Schütze2016, Sprouse2007ContinuousAC}, and we base our annotation methodology on the same.



\begin{figure}[!b]
    \centering
    \includegraphics[width=1\linewidth]{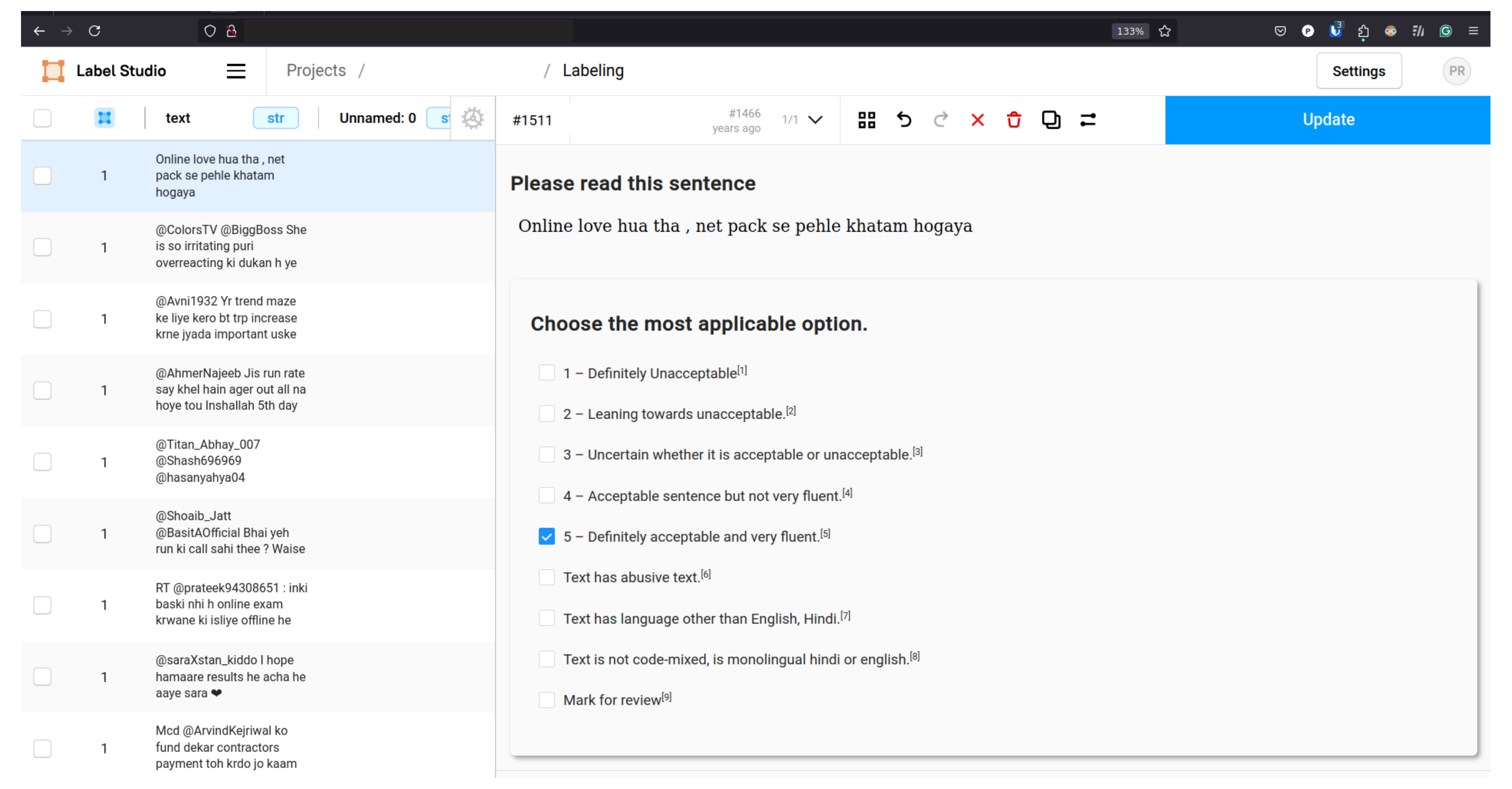}
    \caption{Screenshot of the annotation tool UI provided to the annotators}
    \label{fig:annotion-tool}
\end{figure}

For collecting annotations, we present a code-mixed sample and ask the annotator to rate a sentence on a scale of 1 to 5. In order to have a common grounding of this scale across all our annotators, we describe each label in natural language. We use additional labels to exclude any abusive, monolingual, or other language text from the dataset. These additional labels are presented because some of the samples are taken from OSN and unwanted samples might trickle through despite our manual inspections. Figure~\hbox{\ref{fig:annotion-tool}} shows the user interface used for collecting annotations. The annotation exercise was conducted using LabelStudio.\footnote{https://labelstud.io/ We provide our annotators with a couple more guidelines: a) Annotators should judge a sentence in its entirety and not partially; b) for romanised Hindi words, ``correct spelling'' should not be a factor in rating a sentence lower/higher since there is no canonical spelling for a romanised Hindi word. }

To build a robust dataset for acceptability, we worked with a volunteer pool of undergraduate students, who spoke Hindi as their mother tongue and studied both Hindi and English until 10th Grade, and each sample was presented to 3 unique annotators. Our annotator pool consisted of annotators coming from diverse parts of the country. While it is noted that acceptability judgements for code-mixed text vary across geographies, having multiple annotators and choosing only high agreement samples ensure that our dataset is representative and robust. We conducted a pilot study wherein 30 annotators (including a subset of the authors) annotated 500 samples, with each annotator evaluating 50 samples. Samples with low agreement were selected and 5-10 samples were re-annotated by the annotators to understand their rating rationale. Annotators were interviewed about task ambiguity, guideline efficacy, improvements to the guidelines, and time taken for annotation. The annotation setup and guidelines were refined based on findings of the pilot study.

To ensure consistency and minimize subjectivity, we implemented several measures throughout the annotation pipeline:
\begin{itemize}
    
\item Annotation Guidelines: We provided annotators with detailed, context-specific guidelines and examples to standardize the annotation criteria. These guidelines were developed iteratively with input from experts and refined based on pilot annotations to enhance clarity and reduce subjective interpretation.

\item Inter-Annotator Agreement (IAA) Monitoring: We continuously monitored IAA, assessing agreement on both initial and revised rounds of annotations. Instances with low IAA were identified and re-evaluated, and further training sessions were conducted with annotators to align their understanding.

\item Pruning high disagreement samples: In cases where significant disagreements arose, we conducted an adjudication step where samples with high disagreement were removed. This additional layer helped to ensure that the final labels reflect a consistent interpretation across annotators.

\item Learning from disagreements : In the field of NLP, there is a growing interest in understanding and encoding ambiguities and subjective interpretations, particularly within highly subjective tasks like Hate/Toxic/Bias Detection\hbox{~\cite{10.1613/jair.1.12752}}. Rather than viewing subjectivity as a limitation, we see it as a valuable characteristic that reflects the complex, nuanced nature of code-mixing. By collecting and retaining these subjective variations, our dataset serves as a rich resource for exploring these ambiguities, supporting research on how models can learn to handle and even benefit from subjective diversity. This approach aligns with recent efforts in NLP to create models that are more robust and inclusive, adapting to a broader range of linguistic and human perceptions of code-mixing.

\end{itemize}

\subsection{Dataset Overview}
To ensure consistency in our crowd-sourced annotations, we analyse the inter-annotator agreement of our dataset. Intra-Class Correlation~(ICC)~\cite{Shrout1979IntraclassCU, Fisher1992} is a popular family of estimators for both test-retest reliability and inter-rater consistency. There are different forms of ICC estimators, each of which can involve distinct assumptions and can therefore lead to very different interpretations~\cite{Koo2016AGO, Shrout1979IntraclassCU}. ICC1k is a suitable metric for assessing agreement when: a) reliability of the average ratings is being assessed, and b) when annotators are chosen randomly from a larger pool of annotators. We use ICC1k for assessing the reliability of mean ratings as the aforementioned assumptions hold good for our annotation methodology~\cite{wong-paritosh-2022-k}. ICC scores that are in the range of 0.75 to 0.9 indicate good reliability~\cite{Koo2016AGO}. 

\input{sections/testtable}
\input{tables/02-example-annotations}
As can be seen from Table~\ref{tab:IAA}, for both GCM and OSN data, ICC1k values indicate good reliability for samples with a sum of pair-wise differences less than 4. Sum of pair-wise differences captures the extent of disagreement, and is computed by summing up the absolute difference between all pairs of annotators. For instance, if a sample was annotated as 2,3,4 by the three annotators, sum of pair-wise differences is computed as |2-3| + |2-4| + |3-4| = 4. We use Sum of pair-wise differences as our Disagreement scores, and both terms are used interchangeably. Sum of pair-wise differences will be the range of 0 to 8, since the annotations are on the integers in the range of 1 to 5. For further analysis and experiments we remove samples where disagreement is more than 4. Table~\ref{tab:dataset-stats} shows the total number of samples in our dataset across two sources - Social Network and Synthetically generated sentences. To the best of our knowledge, this is the largest dataset containing annotations for human acceptability judgements for code-mixed sentences. Table~\ref{tab:dataset_examples} shows demonstrative examples for different bins of average rating and agreement between annotators.

\begin{figure*}[!ht]
\includegraphics[width=\textwidth]{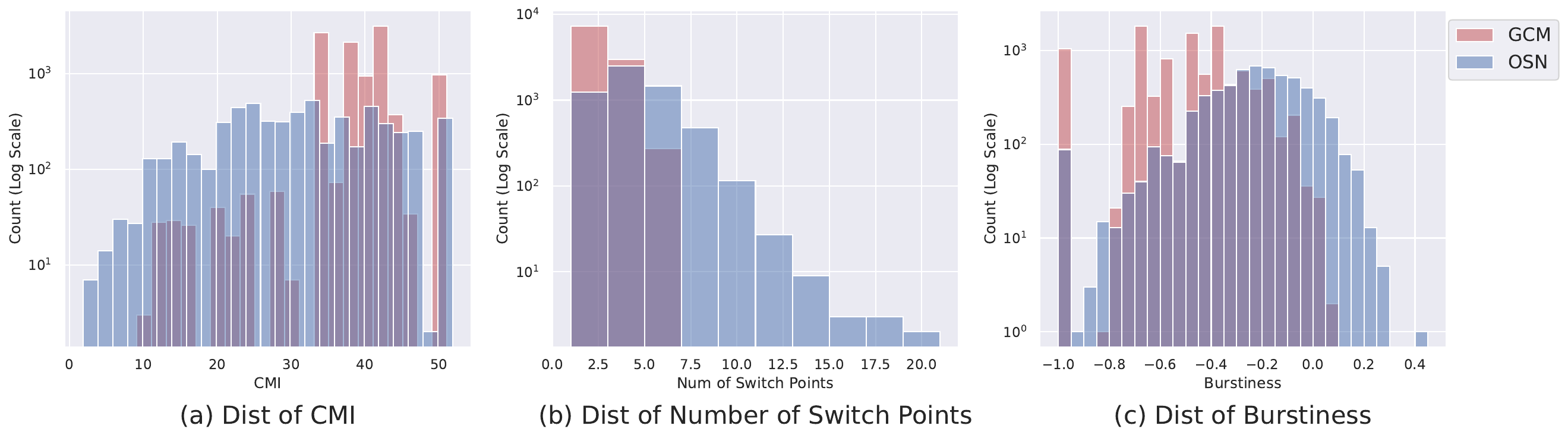}
\caption{Distribution of Code-Mix Index~(CMI), Number of switch points, and Burstiness of code-mix sentences in our dataset. }
\label{fig:appendix_samples_metrics_distribution}
\end{figure*}

\begin{figure*}[!ht]
\includegraphics[width=\textwidth]{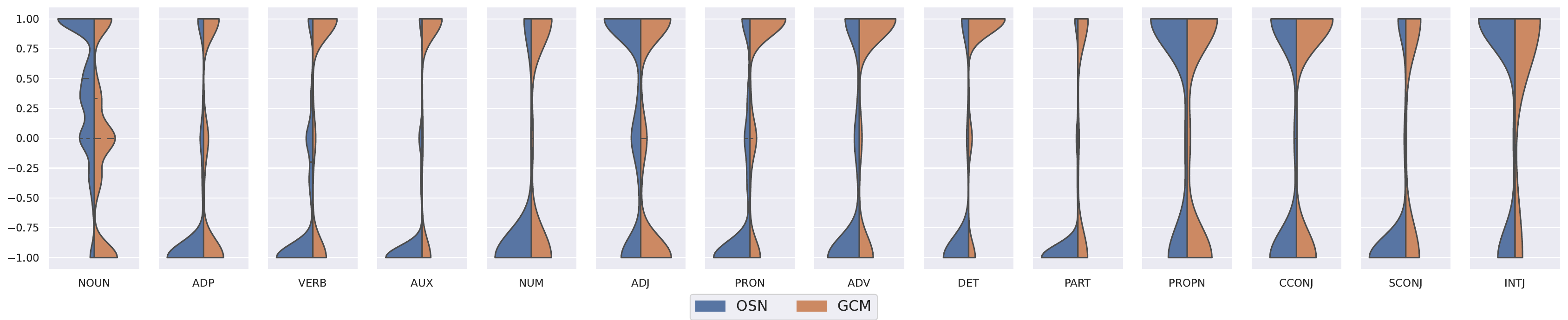}
\caption{Distribution of $SyMCoM$ for OSN and GCM samples in our dataset. $SyMCoM$ score represents the syntactic variety in a code-mix sentence. Distribution of SyMCoM for OSN posts is usually skewed towards -1~(i.e only one language is contributing tokens for that PoS tag). SyMCoM for GCM is spread out for most PoS tags, indicating higher diversity in switching patterns of PoS tags.}
\label{fig:appendix_dataset_distributions}
\end{figure*}


The diversity of the dataset is demonstrated through the distributions of three key metrics—Code-Mix Index (CMI), the number of switch points, and burstiness—illustrated in Figure~\ref{fig:appendix_samples_metrics_distribution}. These metrics collectively capture essential characteristics of code-mixed text. The distribution of CMI (\hbox{Figure~\ref{fig:appendix_samples_metrics_distribution}(a))} reveals a wide range of code-mixing levels, from minimally to highly mixed sentences. The variation in the number of switch points \hbox{(Figure~\ref{fig:appendix_samples_metrics_distribution}(b))} encompasses sentences with both few and frequent alternations between languages. Similarly, the burstiness distribution \hbox{(Figure~\ref{fig:appendix_samples_metrics_distribution}(c))} reflects diverse temporal patterns in language alternation. Compared to GCM, OSN samples have longer tail in the distribution of switch points, and this can be attributed to very long OSN posts. CMI, Number of Switch Points and Burstiness of samples of both OSN and GCM subsets are spread across the spectrum, demonstrating the diversity in the dataset.

SyMCoM is a metric ranging from [-1,1], where the extremities correspond to a particular language. For example, if $SyMCoM_{PoS}$ has peak only at -1, it denotes that that particular PoS tag is contributed by only one language~(Hindi), and a peak at zero indicates that both languages have contributed equally for tokens of that particular PoS tag. In Figure~\ref{fig:appendix_dataset_distributions}, SyMCoM scores for each PoS tag are contrasted for OSN and GCM sets in our dataset. Here, SyMCoM scores for GCM are spread across two peaks, where as OSN peak at only one extreme, indicating a higher variety in patterns of code-mixing present in GCM set. The distributions show peaks at both ends and substantial presence in the mid-range, reflecting a wide spectrum of syntactic mixing, from monolingual dominance to balanced language contributions. Notably, the broader spread of SyMCoM for GCM samples indicates higher syntactic diversity across PoS tags. This observation lends support to our argument for including both human generated and synthetically generated code-mixed text to ensure diversity of code-mixing patterns in datasets. 

\begin{wrapfigure}{r}{0.5\textwidth}
    \begin{center}
        \includegraphics[width=0.42\textwidth,trim={0.9cm 0.65cm 2.4cm 2cm},clip]{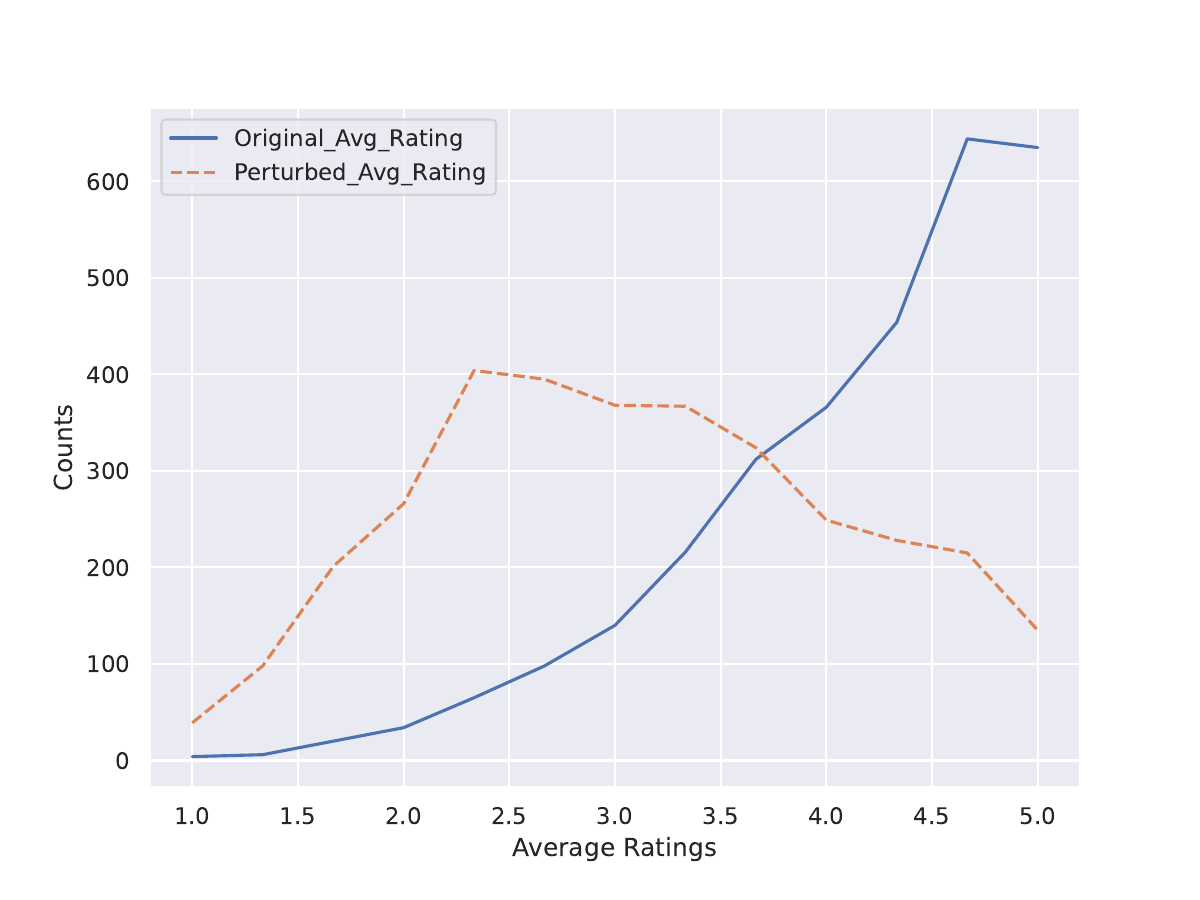}
    \end{center}
    \caption{Impact of perturbations on average ratings of OSN samples. Average ratings of perturbed OSN samples is skewed towards lower acceptability ratings, validating our approach of using perturbations to create negative samples for OSN data.}
\label{tab:appendix_impact_of_perturbtions}
\end{wrapfigure}

As described in Section~\ref{sec:03_data_OSN}, we used perturbations to create OSN samples that are on the lower side of acceptability scale. Figure~\ref{tab:appendix_impact_of_perturbtions} shows the distribution of average ratings for unperturbed and perturbed OSN samples, clearly demonstrating that our perturbations result in creating samples that are on the unacceptable end of the rating scale. Without these perturbations, OSN samples would be heavily skewed towards the higher end of the acceptability scale.

\textbf{Dataset Diversity and Representativeness}: The dataset's dual-source composition further enhances its diversity and representativeness. Combination of synthetic and user generated code-mixed sentences captures both grammatically sound instances and the variability and noise typical of social network text, such as informal language, misspellings, and unconventional structures. Additionally, the dataset comprises approximately 16,642 samples, each annotated by three independent annotators to ensure reliability. Annotations include the original code-mixed sentences, their normalized forms, and romanized versions, enabling a comprehensive analysis of various aspects of code-mixed text. The breadth of sources, meticulous annotations, and robust design make the dataset a valuable resource for studying code-mixing phenomena across diverse contexts.

\begin{figure*}[!h]
    \centering
    \begin{subfigure}{0.45\textwidth}
        \includegraphics[width=\textwidth]{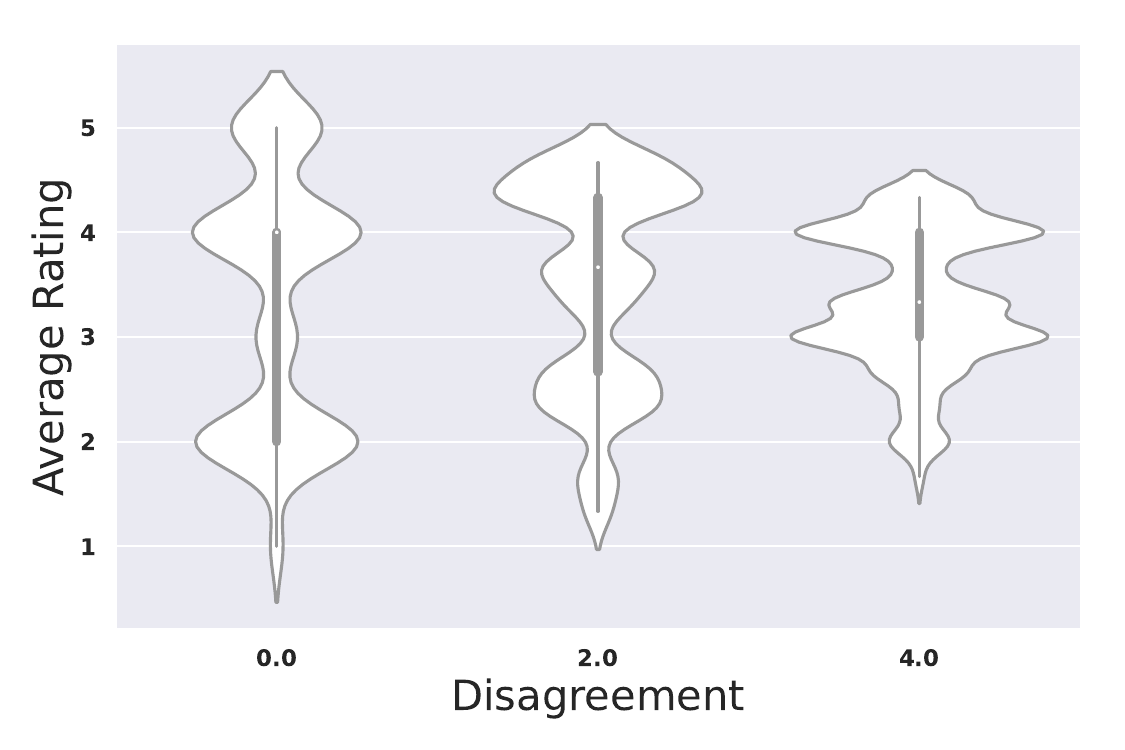}
        \caption{GCM}
        \label{fig:annotations-dist-gcm}
    \end{subfigure}
    \begin{subfigure}{0.45\textwidth}
        \includegraphics[width=\textwidth]{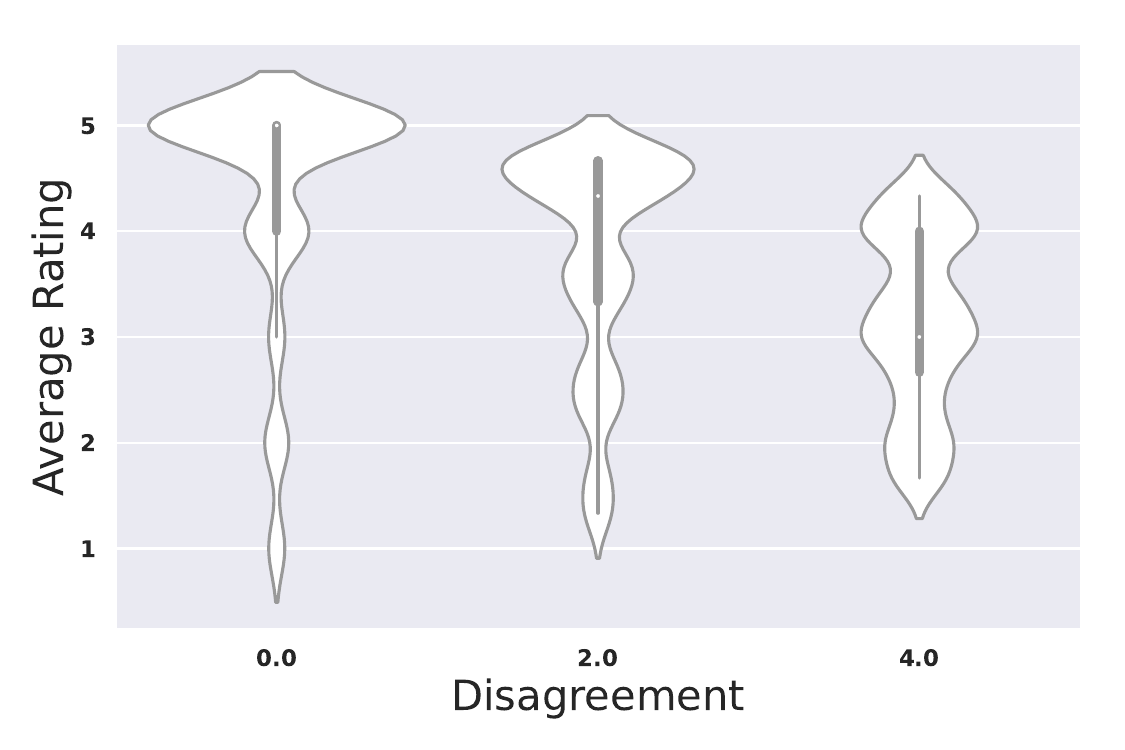}
        \caption{OSN}
        \label{fig:annotations-dist-osn}
    \end{subfigure}
\caption{Distribution of Average Ratings and Disagreement Scores. (a) For GCM average ratings are distributed uniformly, with a large number of samples having high agreement and spread across different ratings. (b) In contrast, for OSN, the majority of samples are rated towards the acceptable side.}
\label{fig:annotations-dist}
\end{figure*}

\textbf{Distribution of Human Ratings}
Figure~\ref{fig:annotations-dist} shows the distribution of collected human ratings for different levels of disagreements. As each sample in the dataset has three annotators, we condense the three ratings to a) the Average Rating, and b) the Sum of pair-wise differences~(Disagreement). The sum of pair-wise difference is indicative of disagreement between the three annotators. 

As seen in Figure ~\ref{fig:annotations-dist}, ratings for GCM are spread out more evenly, with high-agreement samples spread out across the rating scale. Whereas for OSN data, most of the samples are rated towards the acceptable side of the rating scale. Since OSN samples are human-generated, they are expected to be natural, and for negative samples we used perturbations (Sec. \ref{sec:03_data_OSN}). Most of the unperturbed samples are rated higher (acceptable), and perturbed samples shift the distribution of average ratings, giving us samples across the scale. However, even with perturbed samples, OSN data is skewed towards the acceptable end of the rating scale. Disagreement scores show a similar pattern for perturbed/unperturbed posts. 
%


\begin{figure*}[!b]
    \centering
    \begin{subfigure}{0.9\textwidth}
        \includegraphics[width=\textwidth]{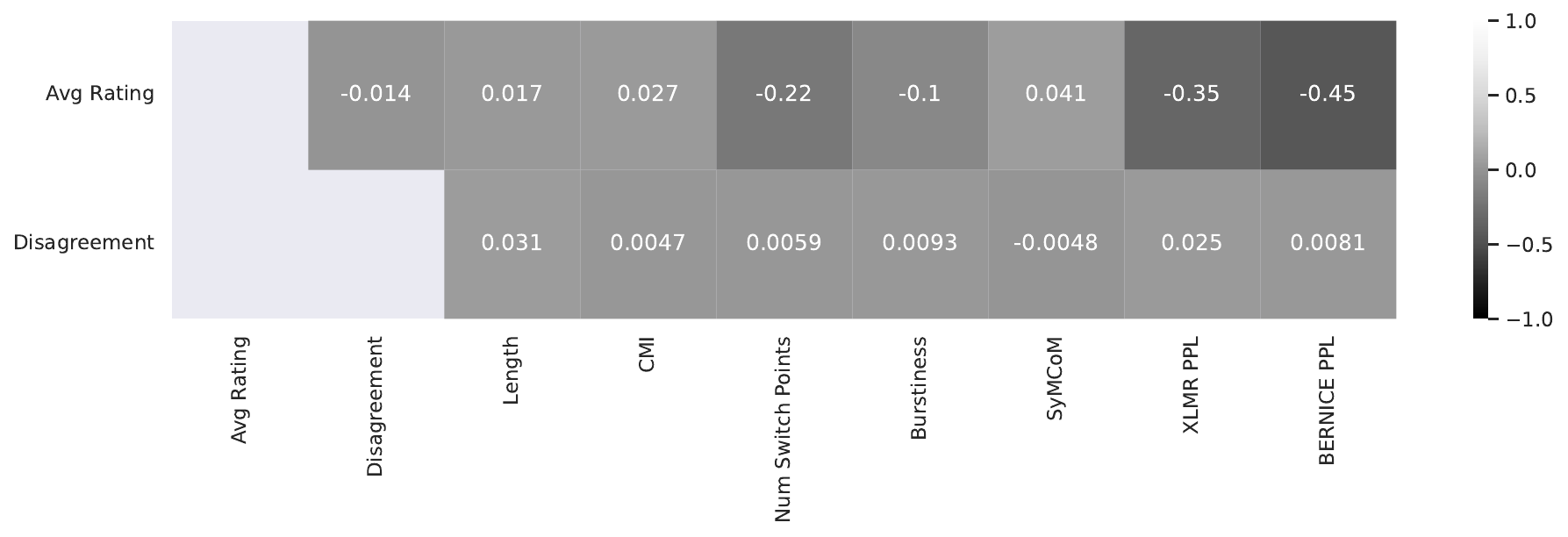}
        \caption{GCM}
        \label{fig:correlation-gcm}
    \end{subfigure}
    \begin{subfigure}{0.9\textwidth}
        \includegraphics[width=\textwidth]{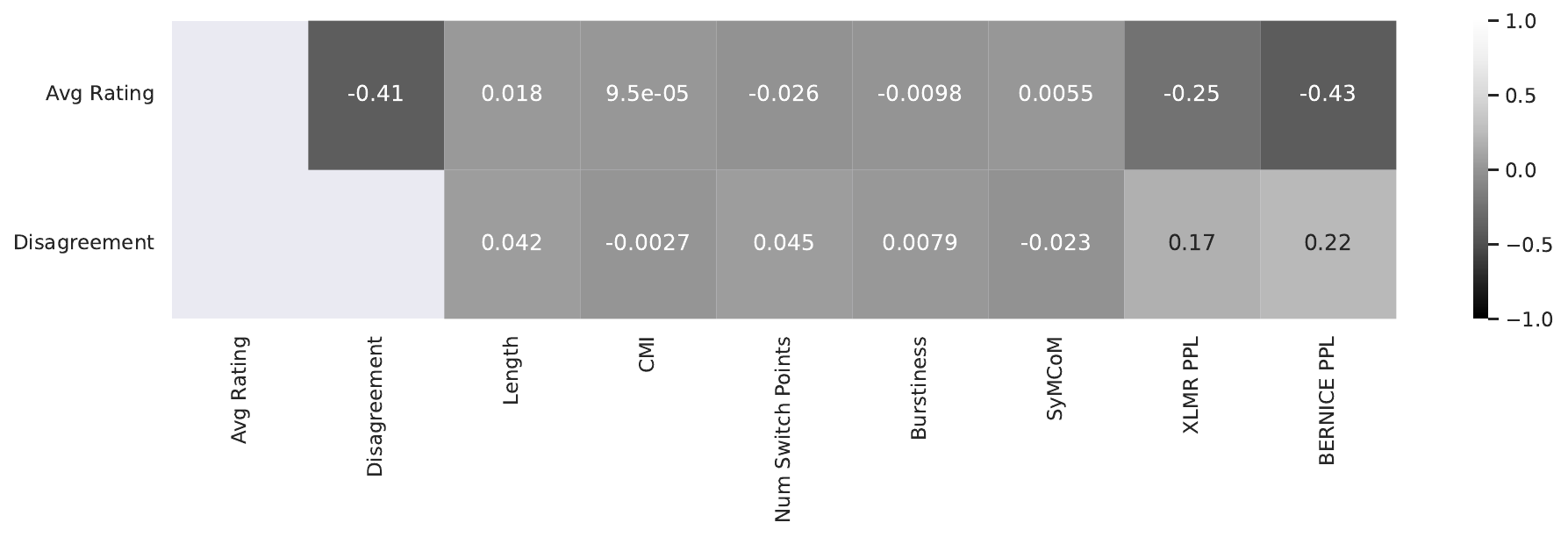}
        \caption{OSN}
        \label{fig:correlation-osn}
    \end{subfigure}
  \caption{Correlation between human acceptability rating and code-mix metrics. For both GCM and OSN, code-mix metrics do not show very low correlations with human acceptability ratings. MLLMs perplexity scores, however, show comparatively high correlation with average ratings.}
  \label{fig:correlations}
\end{figure*}
We also observed that for annotating OSN samples our annotators took thrice the time they did to annotate GCM samples\footnote{Time taken to annotate samples were recorded by the annotation tool.}~(see Table \ref{tab:dataset-stats}). Increased time to annotate OSN samples can possibly be attributed to a) social media text being different from standard forms of writing and; b) perturbations introduced by us. Because humans require more time to evaluate OSN samples compared to synthetically generated code-mixed text, we hypothesise that models predicting the acceptability of OSN code-mixed text would also underperform when compared to GCM samples.

%% file: sections/testtable.tex
    \begin{table}[t]
        \begin{minipage}[t]{0.48\textwidth}

            \raggedright
            \caption{Inter-Annotator Agreement (ICC1k) for GCM and OSN data. Samples where the Sum of Absolute Pair-wise Differences is $\leq4$ have good agreement (ICC1k between 0.7 and 0.9).}
            \label{tab:IAA}

            \resizebox{\linewidth}{!}{%
            \begin{tblr}{
              cells = {c},
              cell{1}{1} = {r=2}{},
              cell{1}{2} = {c=2}{},
              cell{1}{4} = {c=2}{},
              hline{1,8} = {-}{0.08em},
              hline{2} = {2-5}{0.03em},
              hline{3} = {-}{0.05em},
            }
            {Disagreeement} & GCM   &                           & OSN   &                           \\
                                             & ICC1k & {\% of samples \\covered} & ICC1k & {\% of samples \\covered} \\
            0                                & 1     & 16\%                      & 1     & 23\%                      \\
            0-2                              & 0.92  & 58\%                      & 0.93  & 65\%                      \\
            0-4                              & 0.79  & 87\%                      & 0.85  & 87\%                      \\
            0-6                              & 0.68  & 98\%                      & 0.78  & 97\%                      \\
            0-8                              & 0.66  & 100\%                     & 0.75  & 100\%                     
            \end{tblr}
            }

        \end{minipage}%
        \quad
        \begin{minipage}[t]{0.45\textwidth}
            \raggedleft
            \caption{Dataset Statistics. On average, annotators took $\approx 3$x more time to annotate an OSN sample when compared to a GCM sample.}
            \label{tab:dataset-stats}            
            \resizebox{\linewidth}{!}{%

            \begin{tabular}[t]{ccc} 
            \hline
            \multicolumn{1}{l}{}                                                       & \begin{tabular}[c]{@{}c@{}}\textbf{Number of }\\\textbf{Samples}\end{tabular} & \begin{tabular}[c]{@{}c@{}}\textbf{Mean Annotation }\\\textbf{Time per sample}\end{tabular}  \\ 
            \hline
            \begin{tabular}[c]{@{}c@{}}Synthetic Code Mix \\ Text (GCM)\end{tabular}    & 10,726                                                                        & 29.51 seconds                                                                                \\ 
            \hline
            \begin{tabular}[c]{@{}c@{}}Online Social Network \\ Text (OSN)\end{tabular} & 5,916                                                                         & 91.20 seconds                                                                                \\
            \hline
            \end{tabular}
            }

        \end{minipage}%
    \end{table}

%% file: tables/02-example-annotations.tex
\begin{table}[t]
\caption{Demonstrative examples from our dataset, Cline. Examples show the text with romanised Hindi tokens. Full agreement samples are where all the three annotators agreed, while the samples in disagreement rows are the ones where three annotators disagree. For both categories, we present samples where the rating is low (in the range [0-3]), and where the average rating is high (in the range [4-5]). For OSN posts, the spellings, punctuations are same as original text, with user mentions, hashtags and URLs replaced with respective placeholders.}
\label{tab:dataset_examples}
\resizebox{0.9\linewidth}{!}{%
\begin{tabular}{ccc} 
\hline\hline
\textbf{Agreement }                                                       & \begin{tabular}[c]{@{}c@{}}\textbf{Average }\\\textbf{Acceptability Rating}\end{tabular} & \textbf{GCM}                                                                                                                                                              \\ 
\hline\hline
\multirow{2}{*}{\begin{tabular}[c]{@{}c@{}}Full \\Agreement\end{tabular}} & \begin{tabular}[c]{@{}c@{}}Low\\0-3\end{tabular}                                         & \begin{tabular}[c]{@{}c@{}}Whatever may be the problem , atmahatya is nahi koi .\\The final selection will be kiya rajya sarkar ke se.\end{tabular}                       \\ 
\cmidrule{2-3}
                                                                          & \begin{tabular}[c]{@{}c@{}}High\\3-5\end{tabular}                                        & \begin{tabular}[c]{@{}c@{}}last year first National Agriculture Policy ki ghoshna ki gayhi thi\\Thereafter , the body has been sent postmortem ke liye .\end{tabular}     \\ 
\hline\hline
\multirow{2}{*}{Disagreement}                                             & \begin{tabular}[c]{@{}c@{}}Low\\0-3\end{tabular}                                         & \begin{tabular}[c]{@{}c@{}}They have sustained serious injuries on chaahti aur siir gahri .\\It said that person was pronounced dam thod ghatnasthal par .\end{tabular}   \\ 
\cmidrule{2-3}
                                                                          & \begin{tabular}[c]{@{}c@{}}High\\3-4\end{tabular}                                        & \begin{tabular}[c]{@{}c@{}}He was immediately shifted to nearby aspataal upchaar ke liye. \\in Nagaland , Mizoram and Tripura bhi yahi haal hai .\end{tabular}            \\ 
\hline\hline
\multicolumn{1}{l}{}                                                      & \multicolumn{1}{l}{}                                                                     & \textbf{OSN}                                                                                                                                                              \\ 
\hline
\multirow{2}{*}{\begin{tabular}[c]{@{}c@{}}Full \\Agreement\end{tabular}} & \begin{tabular}[c]{@{}c@{}}Low\\0-3\end{tabular}                                         & \begin{tabular}[c]{@{}c@{}}Sir rajnigandha mai ke bare ditels me janna chahta hu give me\\vah emoji itnaa torture vah hissaa khud karo jaye emoji URL\end{tabular}        \\ 
\cline{2-3}
                                                                          & \begin{tabular}[c]{@{}c@{}}High\\3-5\end{tabular}                                        & \begin{tabular}[c]{@{}c@{}}userMention userMention Hn Na sir kindly duaoon mn yaad rakhna \\userMention No need yaar bas kaam mat chodo\end{tabular}                      \\ 
\hline
\multirow{2}{*}{Disagreement}                                             & \begin{tabular}[c]{@{}c@{}}Low\\0-3\end{tabular}                                         & \begin{tabular}[c]{@{}c@{}}Khud ko dekhe itni badi chale mistake go to the bench \\userMention Kuxh bahut zyada nahin aakar but he nice tweet dear .\end{tabular}         \\ 
\cline{2-3}
                                                                          & \begin{tabular}[c]{@{}c@{}}High\\3-5\end{tabular}                                        & \begin{tabular}[c]{@{}c@{}}RT userMention : userMention Fans hain hum sabse best \#3YearsOfSidNaaz\\Merit se pehle General walo ko hi history bana dega URL\end{tabular}  \\ 
\hline
\multicolumn{1}{l}{}                                                      & \multicolumn{1}{l}{}                                                                     & \multicolumn{1}{l}{}                                                                                                                                                      \\
\multicolumn{1}{l}{}                                                      & \multicolumn{1}{l}{}                                                                     & \multicolumn{1}{l}{}                                                                                                                                                     
\end{tabular}
}

\end{table}

%% file: sections/04-dataset-Analysis.tex
\section{Analysis}\label{sec:04_dataset_analysis}
We analyse our dataset to explore the relationship between acceptability judgements, code-mix metrics, and the pseudo-perplexity of multilingual large language models.



\textbf{Relationship between acceptability judgements, code-mixing metrics and MLLMs Perplexity}
Code-mixing metrics~(see Section~\ref{02_sec:bacground-acceptability-codemix}) are computational tools employed for quantifying and comparing the extent of code-mixing in sentences or corpora. These metrics capture lexical and syntactical properties of switching. However, their capability to evaluate the naturalness or acceptability of code-mixed sentences has not been evaluated. To assess the relationship between code-mixing metrics and the human acceptability ratings in our dataset, we compute correlations between human ratings and multiple code-mix metrics - Code-Mixing Index~\citep{gamback-das-2016-comparing}, Burstiness~\citep{guzman17_interspeech}, Number of Switch Points. 

Traditionally, a language model's perplexity has been used to assess likelihood of a sentence. Encoder-only LLMs' surprisal scores~(or pseudo-log-likelihood scores) have been used for ranking sentences, and also for assessing acceptability of English sentences~\cite{salazar-etal-2020-masked}. With recent adoption of Multilingual LLMs in code-mixed settings, it becomes pertinent to analyse MLLMs familiarity with the notion of ``acceptable'' code-mixed text. To test this, we compute correlation between surprisal scores~\citep{salazar-etal-2020-masked} of Multilingual Large language models~(XLM-R and Bernice) and acceptability ratings of code-mixed text. Correlations are shown in Figure~\ref{fig:correlations}. 

For GCM data, average ratings show the highest correlation with the Number of Switch Points~(-0.2), implying that as the Number of Switch Points in a sentence increase, the average ratings decrease. However, such a pattern is not observed for OSN samples, with all the correlations being very close to zero. Disagreement scores, however, show no significant correlation with any of the code-mixing metrics. Furthermore, MLLMs negative pseudo log likelihood scores are weakly correlated~(0.2<$\mid\rho\mid$<0.5) with average acceptability ratings. However, MLLM's surprisal scores have better correlation with acceptability ratings as compared to code-mixing metrics, providing indirect evidence of MLLMs utility in differentiating between acceptable and unacceptable code-mixed text.

Traditional code-mixing metrics are primarily focused on capturing surface-level characteristics of code-mixed text, such as the frequency of switches or the structural arrangement of words from different languages. While these metrics provide valuable quantitative information, they do not capture deeper linguistic and contextual nuances that contribute to the naturalness or acceptability of a code-mixed sentence. 
To draw an analogy, by only analyzing the POS tags of a monolingual sentence may reveal its syntactic structure but does not provide the complete picture into how natural or fluent the sentence sounds. Similarly, code-mixing metrics fail to account for the intricate interplay of syntax, semantics, pragmatics, and sociolinguistic factors that influence human judgments of code-mixed text acceptability. These deeper linguistic properties are precisely what the traditional metrics are missing. We also refer to Table \hbox{\ref{tab:examples_codemixmetrics}}, where we present examples of code-mixed sentences along with their corresponding code-mix metric values and human acceptability scores. These examples illustrates that sentences with similar metric values can vary significantly in their acceptability, demonstrating the limitations of these metrics in capturing the qualitative aspects of code-mixed text. Even when the metric values differ slightly, they remain largely insensitive to the subtle variations that influence human perception.

\textbf{Regression Model} For more comprehensive analysis, we perform a (linear) regression analysis between code-mixing metrics and acceptability ratings. While correlation analysis gives a sense of the strength and direction of the linear relationship between two variables, linear regression provides a more detailed understanding by showing relative predictive capabilities of independent variables. Despite limited individual correlations between code-mixing metrics and acceptability ratings, regression analysis becomes essential to assess if a linear combination of these metrics can effectively explain code-mixing acceptability. Regression coefficients were computed after all the independent variables were normalised to lie between [0, 1], allowing us to compare coefficients from the model. Regression coefficients are shown in Table~\ref{tab:regression_coeff}. 

\input{tables/06-regression-coeff}

"Intercept" term (3.2*** for GCM and 3.64*** for OSN) represents the baseline average rating when all other variables are held constant. The significance of the intercept highlights that, even in the absence of specific code-mixing features, there is a baseline rating trend in the data. This intercept captures the inherent average rating in the dataset, providing a reference point against which the effects of other variables are measured. The metrics "Number of Switch Points" and "$SyMCoM_{sent}$" show the strongest effects on acceptability ratings, with significant coefficients across both datasets (GCM and OSN). The regression analysis includes $SyMCoM$ metrics, which captures syntactic complexities in code-mixed text. This ensures that syntactic properties are accounted for in the modeling process. Specifically, a higher "Number of Switch Points" negatively impacts ratings, with coefficients of -1.12 and -0.57 for GCM and OSN, respectively. This suggests that frequent language switches reduce the perceived acceptability of code-mixed content. Conversely, the positive coefficients for "$SyMCoM_{sent}$" (0.40 for GCM and 0.34 for OSN) imply that greater syntactic similarity contributes positively to ratings, potentially enhancing readability and coherence in code-mixed utterances. The "Length" metric has a statistically significant positive effect on ratings in both datasets, with coefficients of 0.32 (GCM) and 0.52 (OSN). This indicates that longer utterances tend to be rated more favorably, possibly because they allow for richer context and more coherent language blending. This finding might reflect a preference for contextually meaningful or complete statements over shorter, fragmented code-mixed expressions. The "CMI" metric shows minimal impact on acceptability ratings, with non-significant coefficients in both datasets (0.03 and 0.05). This suggests that the overall degree of code-mixing, as measured by CMI, is not a strong determinant of rating. Similarly, "Burstiness" has a relatively weak and inconsistent effect, with a significant negative coefficient in GCM (-0.12) but not in OSN. This could indicate that sudden bursts of language switching might disrupt coherence in some contexts, but not universally across datasets. Despite the significant relationships identified, the low R² value (< 0.1) indicates that while these variables have some influence, the overall variance in acceptability ratings is not well-explained by the code-mixing metrics alone. This highlights the complexity of acceptability judgments, which may be influenced by factors beyond the scope of the current metrics, such as semantic coherence or cultural familiarity with code-mixing. The low R² value suggests that acceptability judgments are complex and multifaceted, and future research should explore additional factors—such as semantic coherence, cultural familiarity, or individual annotator biases—to gain a more comprehensive understanding.

In summary, our analysis reveals that most code-mixing metrics have no correlation with acceptability ratings of code-mixed text, with Number of Switch Points being the only exception. MLLMs' surprisal scores show better correlation with acceptability ratings than code-mixing metrics, indicating the utility of MLLMs in distinguishing acceptable and unacceptable code-mixed text. Regression analysis underscores the limited explanatory power of code-mixing metrics for acceptability ratings, with Number of Switch Points and SyMCoM being more impactful compared to CMI and Burstiness, for both GCM and OSN samples. Notably, CMI has a statistically insignificant impact on acceptability ratings, offering valuable insights for future research.




    

%% file: tables/06-regression-coeff.tex

\begin{table}[!t]
\caption{Regression coefficients obtained by regressing average rating against code-mixing metrics. $R^2$ is non-zero but very low~(<0.1). Number of Switch Points and SyMCoM have relatively higher impact on acceptability ratings. *** indicates p-value < 0.005, ** indicates p-values < 0.05, * indicates p-value < 0.1.}
\label{tab:regression_coeff}
\centering
\resizebox{0.4\linewidth}{!}{%
\begin{tabular}{lcc} 
\hline\hline
\multicolumn{1}{c}{\multirow{2}{*}{\begin{tabular}[c]{@{}c@{}}Independent~\\Variable\end{tabular}}} & \multicolumn{2}{c}{Average Rating}                         \\ 
\cline{2-3}
\multicolumn{1}{c}{}                                                                                & GCM                         & OSN                          \\ 
\midrule
CMI                                                                                                 & 0.03                        & 0.05                         \\ 
\midrule
Num. of Switch Points                                                                               & -1.12***                    & -0.57***                     \\ 
\midrule
Burstiness                                                                                          & -0.12**                     & -0.11                        \\ 
\midrule
$SyMCoM_{sent}$                                                                                       & 0.40***                     & 0.34**                       \\ 
\midrule
Length                                                                                              & \multicolumn{1}{l}{0.32***} & \multicolumn{1}{l}{0.52***}  \\ 
\midrule
Intercept                                                                                           & 3.2***                      & 3.64***                      \\
\hline\hline
\end{tabular}
}
\
\end{table}

%% file: sections/05-classifier.tex
\section{Predicting Acceptability}\label{sec:05_classifier}
We train several models to predict acceptability ratings on Cline, covering all data settings. We pose the acceptability prediction task as a regression task - given a code-mixed sentence predict the rating on a scale of 1 to 5. We split our dataset into train-dev-test in 70:10:20 ratio, while ensuring that distribution of ratings is similar across the splits. We use Root Mean Square Error~(RMSE) and Mean Absolute Error~(MAE) as our evaluation metrics. We carried out experiments using three approaches: a) using only code-mixing metrics; b) fine-tuning multilingual language models; c) zero and few-shot prediction using large language model.

We benchmarked our approaches against two baselines: a) Random Predictions; b) Human Annotations Baseline. Random Predictions are created by picking a random value from a uniform distribution between 1 to 5. For the Human Annotations Baseline, we calculated the RMSE and MAE between each pair of annotators, then averaged the results across all pairs of annotators.

\textbf{Using Code-Mix metrics to predict Acceptability:} In Section~\ref{sec:04_dataset_analysis}, our Correlation and Regression analysis established that a linear combination of code-mix metrics is insufficient to explain acceptability of code-mixed sentences. However, it is possible that using our dataset we can learn a complex non-linear mapping between code-mix metrics and acceptability. Simple Feed-Forward Networks are capable of learning complex non-linear mapping functions. In order to predict acceptability for code-mixed text, we combine the following the code-mix metrics to create feature vectors for our dataset to train a single-layer feed-forward network: a) Sentence Length; b) CMI; b) Burstiness; c) Number of Switch Points; d) Sentence level SyMCoM Score. 

While code-mix metrics on their might not be able to model acceptability ratings of code-mixed text, it is quite possible that when the code-mix metrics are combined with features from pre-trained langauge model can help reliably model acceptability of code-mixed text. To test this hypothesis, we present results from a feed-forward network that incorporates two sets of features: code-mixing metrics and the perplexity values of a multilingual pre-trained language model. These perplexity values serve as a proxy for the semantic properties encoded by the language model. By combining syntactic features (via $SyMCoM$), code-mix features, and semantic proxies (via perplexity), we effectively integrate some semantic and syntactic features for improved modeling. The results in Tables \hbox{\ref{tab:regression_coeff}} and \hbox{\ref{tab:results_seperate_training_withChatGPT}} demonstrate that even when combining syntactic, semantic, and code-mixing features, the complex phenomenon of code-mixed text acceptability remains difficult to model effectively. This highlights the inherent challenges of capturing such a multifaceted sociolinguistic phenomenon using traditional feature-based approaches. Based on our findings, fine-tuning a much larger model directly on the task-specific data appears to be the most promising solution for better capturing these nuances. 






\textbf{Pre-trained Multilingual Large Language Models:}
Motivated by recent application of pre-trained language models on code-mixed tasks~\cite{khanuja-etal-2020-gluecos, aguilar-etal-2020-lince}, we fine-tuned multi-lingual large language models for predicting acceptability scores for Code-mixed Hindi-English text.

We selected nine state-of-the-art models, namely \hbox{XLM-R~\citep{conneau-etal-2020-unsupervised}}, \hbox{Bernice~\citep{delucia-etal-2022-bernice}}, \hbox{IndicBERT~\citep{kakwani2020indicnlpsuite}}, \hbox{mT5~\cite{xue-etal-2021-mt5}}, \hbox{mBART~\cite{liu-etal-2020-multilingual-denoising}}, Llama 3.2 1B and 3B models \footnote{\url{https://huggingface.co/collections/meta-llama/llama-32-66f448ffc8c32f949b04c8cf}}, \hbox{Qwen 2.5 1B model \cite{qwen2.5, qwen2}}, \hbox{Phi-3 3B model~\cite{abdin2024phi3technicalreporthighly}}. Our experiments cover all the language model families - Encoder only, Encoder-Decoder models, Decoder only models. Our model selected was based on following characteristics: a) multilinguality; b) likely exposure to code-mixed text or romanised Hindi; c) training on social media text. For decoder only models, much larger models are avaialble, but we limited our experiments till 3 billion parameter models due to computation constraints. As model capacity grows we expect them to perform better on the task, similar to the trends seen in other NLP tasks.


Furthermore, non-Latin/Roman script languages are frequently romanised in user-generated text. For English-Hindi code-mixed text, romanisation of Hindi is common~(Hindi's native script is Devanagari). The prevalence of romanised text in pre-training corpora of MLLMs is expected to influence downstream task performance. To evaluate this impact, we train separate models for romanised and normalised forms, where Hindi tokens are in the native Devanagari script. MLLMs are fine-tuned with a single layer feed-forward network on top of the model~\citep{wolf-etal-2020-transformers}.

All the pre-trained language models are trained without freezing MLLM layers. Our preliminary experiments indicated that training with freezed MLLM weights were consistently outperformed by training with unfreezed MLLM layers. All our models are optimised for mean square error loss using the Adam optimizer with linear weight decay. Effective effective batch size varies for models, typically 32 for Encoder only models, 16 for Encoder-Decoder models, 8 for Decoder only models. Effective batch size was based on max batch size we could fit on respective GPUs - Encoder-only models were trained on Nvidia GTX 1080 Ti GPUs (12GB VRAM), and rest trained on Nvidia A5000 32GB cards. Decoder only models were trained using mixed precision (bf16). We carried out learning rate search over the range of \{1e-2, 1e-3, 1e-4, 1e-5, 1e-6\} for all models, and chose the learning rate which gave best performance over validation set: we use 1e-5 for Encoder-only models, 1e-4 for Encoder-Decoder models and 1e-6 for Decoder-only modeols. Further training details and scripts can be found in \hbox{\href{https://github.com/prashantkodali/CodeMixAcceptability}{our code repository}}.

\textbf{ChatGPT for Acceptability judgements of Code-mix sentences} With recent popularity of Instruction following Large Language Models, and their zeroshot and fewshot capabilities, prompting LLMs have become an important benchmark to compare against. In our study, we use the commercial ChatGPT-3.5-turbo API to benchmark acceptability judgements for code-mix sentences. We anecdotally tested if ChatGPT was able to define what code-mixing is, and if the model was able to generate code-mixed sentences, and initial observations were positive, and hence the model was chosen as part of our experimental setup. 

We conduct our evaluations by querying ChatGPT under both zero-shot and five-shot settings. In order to obtain consistent and minimally verbose responses, we experimented with several prompt formulations aimed specifically at eliciting only the rating number from the model—without any accompanying explanations or commentary. This was particularly important for standardizing outputs and enabling direct comparisons across different experimental configurations. After multiple rounds of prompt engineering and empirical testing, we finalized a set of prompts that reliably prompted the model to return only the required numerical rating. 

For the five-shot setting, we included five labeled examples in the prompt to guide the model's behavior. These examples were carefully sampled from the training set to ensure there was no overlap or leakage into the test set, thereby preserving the integrity of our evaluation. In each prompt, the placeholder <query-sentence> was dynamically replaced with a sentence drawn from the test sets of two distinct domains in our study: GCM and OSN. We query the API using both forms, the normalised as well as the romanised text, for a thorough comparison across all our experiments. Furthermore, to comprehensively evaluate the model’s robustness and sensitivity to script variation, we issued API queries using both normalized (standardized orthography) and romanized (phonetically transliterated into Roman script) forms of the input sentences.

Following prompts were used in our study.

\begin{tcolorbox}[colback=lightgray,coltext=black]
\begin{center}
    \textbf{Zeroshot Prompt}
\end{center}
``Score the following Hinglish sentences based on how natural they sound with a floating point number on a scale of 1 to 5.  Respond with a single number. Do not explain.\newline 
Sentence : <query-sentence>\newline
Response : ''
\begin{center}
    \textbf{5-shot Prompt}    
\end{center}
``Score the following Hinglish sentences based on how natural they sound with a floating point number on a scale of 1 to 5.  Respond with a single number. Do not explain.\newline
Sentence : <example 1>\newline
Response : <rating>\newline
.\newline
.\newline
Sentence : <example 5>\newline
Response : <rating>\newline
Sentence : <query-sentence>\newline
Response : ''

\end{tcolorbox}

\subsection{Results}
Table~\ref{tab:results_seperate_training_withChatGPT} reports the acceptability prediction performance across multiple data settings and models. The results in Table 6 also allow for a comparison of model performance across different architecture families— Encoder-only, Decoder-only, and Encoder-Decoder models.

\input{tables/03-results_new}

The Human Annotations Baseline reflects the average RMSE and MAE across annotator pairs, capturing variability and consistency rather than serving as an absolute ground truth. It significantly outperforms random predictions (GCM: RMSE 1.05, MAE 0.78 | OSN: RMSE 0.78, MAE 1.47 vs. GCM: RMSE 1.54, MAE 1.25 | OSN: RMSE 1.78, MAE 1.47), demonstrating that human annotations provide structured and meaningful labels.  

The Feed Forward Network trained with Code-Mix Metrics achieves performance close to human annotator disagreement levels (GCM: RMSE 0.97, MAE 0.83 | OSN: RMSE 1.42, MAE 1.01), while fine-tuned models like Llama 3.2 - 3B and Qwen 2.5B surpass the human baseline, suggesting they produce more consistent outputs than individual annotators.  

Encoder-only models generally perform well, with IndicBERT standing out as the best among them. IndicBERT achieves an RMSE of 0.72 (normalized) and 0.88 (romanized) on the GCM dataset, and 1.05 (normalized) and 1.17 (romanized) on the OSN dataset. Among encoder-only models, normalized text provides a clear advantage, as observed in the performance differences between normalized and romanized settings.

The decoder-only models demonstrate strong performance, with Llama 3.2 - 3B achieving the best overall results across all models, with an RMSE of 0.68 and MAE of 0.53 on the GCM dataset (normalized setting), and 0.87 and 0.69, respectively, on the OSN dataset. Qwen and Phi models also perform competitively but consistently trail Llama in both normalized and romanized settings. The decoder-only family benefits from normalization, as shown by the superior performance in the normalized configuration.

Encoder-decoder models perform competitively, with mBART showing notable effectiveness, achieving an RMSE of 0.70 and MAE of 0.54 on the GCM dataset (normalized). Similarly, mBART performs well on the OSN dataset with an RMSE of 0.94 and MAE of 0.70 in the normalized setting. Compared to decoder-only models like Llama, encoder-decoder models tend to lag slightly in performance but remain superior to encoder-only models in some scenarios, particularly in the normalized setting.

Overall, Decoder-only models emerge as the most effective architecture for this task, particularly in the normalized setting. Among the encoder-decoder models, mBART is the most competitive. Encoder-only models, while performing well, exhibit slightly lower performance compared to decoder-only and encoder-decoder counterparts. These findings suggest that the architecture type and text normalization both play critical roles in determining model performance on code-mixed text acceptability tasks.

ChatGPT-3.5 Zero-Shot performs worse than the human baseline (GCM: RMSE 1.22, MAE 0.95 | OSN: RMSE 1.28, MAE 1.20), indicating that prompting alone is insufficient for reaching annotator-level agreement. Few-shot prompting offers slight improvements (GCM: RMSE 1.09, MAE 0.87 | OSN: RMSE 1.17, MAE 1.46) but still lags behind fine-tuned models and human performance.




\textbf{Effect of normalisation on model performance:} Since MLLMs are trained with tokens in their native script~(e.g. in the training corpus of XLM-R, monolingual Hindi written in Devanagari script is likely to be much more prevalent than romanised monolingual Hindi, or code-mixed English-Hindi), model's performance drops on romanised text~\cite{pires-etal-2019-multilingual}. As romanisation doesn't have any canonical spelling, we try two different transliteration engines for GCM samples: IndicTrans\footnote{https://github.com/libindic/indic-trans} and IndicXlit~\cite{madhani-etal-2023-aksharantar}. In Table~\ref{tab:results_seperate_training_withChatGPT} results for GCM normalised runs are average of performance of two separate models trained on different transliterated inputs. We observe that normalisation always improves the acceptability predictions of MLLMs. Using IndicTrans as the normalisation engine usually leads to better performance for acceptability prediction.

\input{tables/tokenfertility_perplexity}

To explain the performance differences between normalized and romanized code-mixed text, we analyzed two metrics: tokenizer fertility and perplexity/surprisal. Tokenizer fertility measures the average number of tokens a model's tokenizer generates per word, while perplexity/surprisal quantifies how surprised a model is by a given sentence, with higher values indicating greater difficulty in processing the text. Lesser token fertility, i.e words are broken into lesser number of subwords, is shown to lead to better downstream task performance\hbox{~\cite{rust-etal-2021-good, ahia-etal-2023-languages, thrush2024improvingpretrainingdatausing}}. Similarly, perplexity/surprisal has also been shown to correlate with performance on downstream tasks\hbox{~\cite{thrush2024improvingpretrainingdatausing}}. Table\hbox{~\ref{tab:tokenizerfertility_perplexity}} compares the tokenizer fertility and the model perplexity across datasets and models. For computing the model perplexity we used Huggingface's evaluate library for Decoder-only models and minicons library for computing suprisal for Encoder-only and Encoder-Decoder models.

\begin{itemize}
    \item Tokenizer Fertility: For models like BERNICE, XLM-Roberta, IndicBERT, mT5, and mBART, normalized sentences exhibit lower tokenizer fertility compared to their romanized counterparts. For example, IndicBERT's tokenizer fertility on GCM data is 1.56 for normalized text and 1.81 for romanized text. This indicates that normalization reduces the number of tokens generated per word, likely simplifying the input for the models and enhancing performance. A similar trend is observed for OSN data, where normalized sentences consistently result in fewer tokens compared to romanized ones. Interestingly, decoder-only models like Llama, Qwen, and Phi exhibit the opposite trend. For these models, tokenizer fertility is higher for normalized text than for romanized text. For instance, Phi 3 - 3B has a tokenizer fertility of 2.8 for normalized GCM text but only 1.6 for romanized GCM text. This suggests that decoder-only tokenizers are less optimized for normalized input, potentially explaining why they do not benefit as much from normalization.

    \item Perplexity/Surprisal: Across all models, normalized text consistently results in lower perplexity compared to romanized text. For example, BERNICE achieves a perplexity of 39.2 on normalized GCM data compared to 70.5 on romanized data, and similar trends are observed for OSN data. Lower perplexity values indicate that models find normalized text easier to process and predict, aligning with the performance improvements observed in normalized settings. Decoder-only models also follow this trend, despite their tokenizer fertility differences. For instance, Phi 3 - 3B exhibits a perplexity of 33.1 on normalized GCM data compared to 2139.2 on romanized text, indicating that normalization aids in reducing uncertainty in the model's predictions.

\end{itemize}

The observed performance differences between normalized and romanized text can be attributed to these two metrics. For encoder and encoder-decoder models, normalization reduces tokenizer fertility and perplexity, making the input more efficient and predictable. Decoder-only models, while benefiting from lower perplexity in normalized text, exhibit higher tokenizer fertility, which may partly offset these gains. These insights underscore the critical role of tokenization and model architecture in determining the impact of script normalization on performance.



\textbf{Effect of data source on model performance} Table~\ref{tab:results_seperate_training_withChatGPT} show that all models, across all settings, perform consistently better on GCM data relative to OSN data. We hypothesise that this is because synthetic GCM data, which is much more linguistically grounded, aligns better with pre-trained MLLMs' encoded cross-lingual knowledge. We observe that even Bernice performs much better on GCM for all settings as compared to OSN data, even though it is specifically trained on OSN (Twitter) data. For example, Bernice trained on romanised code-mixed sentences reports a 1.02 RMSE on OSN data as compared to 0.79 RMSE on GCM data~(see Table~\ref{tab:results_seperate_training_withChatGPT}). This result underscores the complexity involved in dealing with code-mixed data where the domain as well as the linguistic rules of code-mixing, play a role in determining model performance.

We conjecture that the superior performance of models like Llama-3.2 may be attributed, in part, to their training data and size when compared to other model. These models might have been exposed to a larger proportion of code-mixed text or Romanized Hindi during pre-training, giving them an advantage in understanding and processing code-mixed sentences. We would like to note that the performance differences observed between models could stem from various factors, including model architectures, training data, and pre-training objectives. Consequently, drawing definitive conclusions about the reasons behind their performance differences is challenging and cannot be substantiated with quantitative evidence within the scope of this study.


\subsection{Error analysis}

\begin{figure*}[b]
    \centering
    \begin{subfigure}{0.32\textwidth}
        \includegraphics[width=\textwidth]{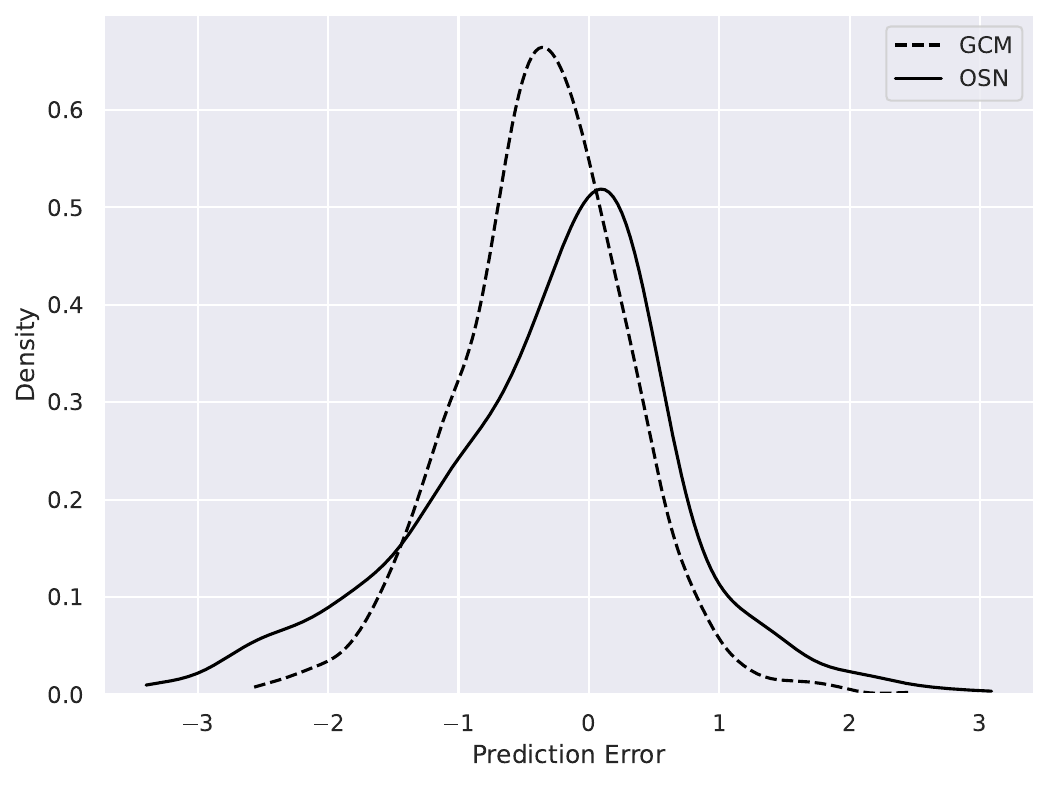}
        \caption{XLM-R Distribution of Errors}
        \label{fig:error-dist-gcm-osn}
    \end{subfigure}
    \begin{subfigure}{0.32\textwidth}
        \includegraphics[width=\textwidth]{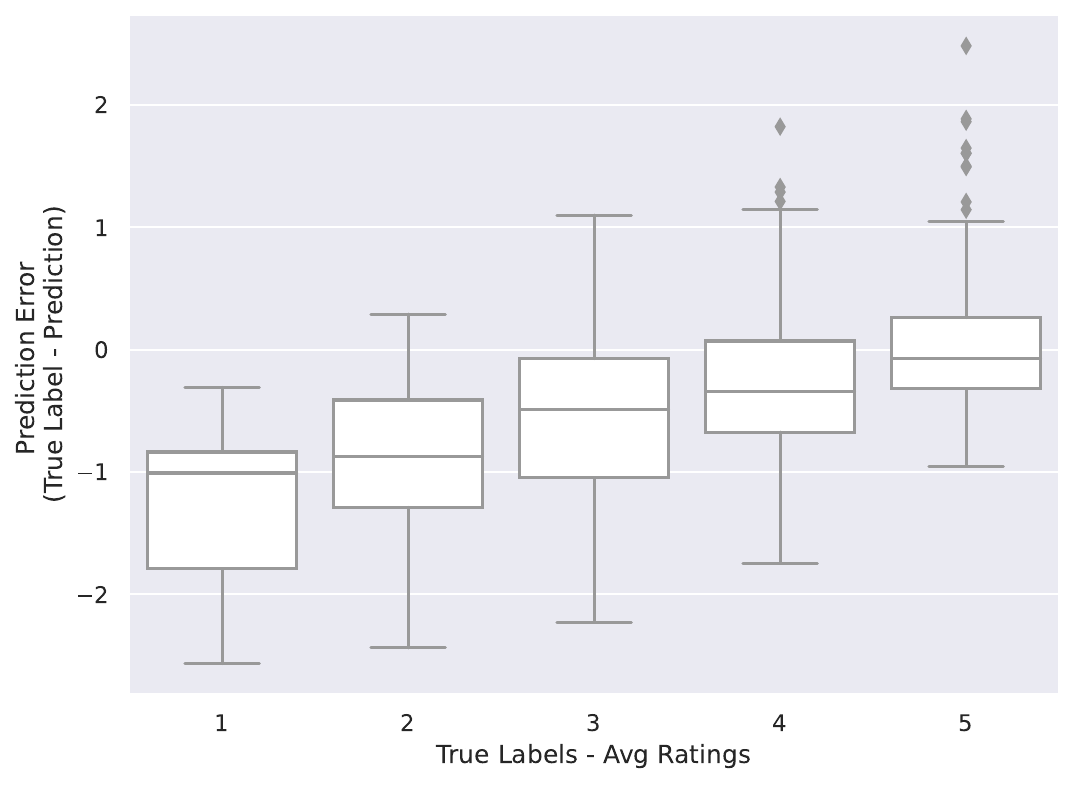}
        \caption{XLM-R GCM Errors}
        \label{fig:errors-prediction-gcm}
    \end{subfigure}
    \begin{subfigure}{0.32\textwidth}
        \includegraphics[width=\textwidth]{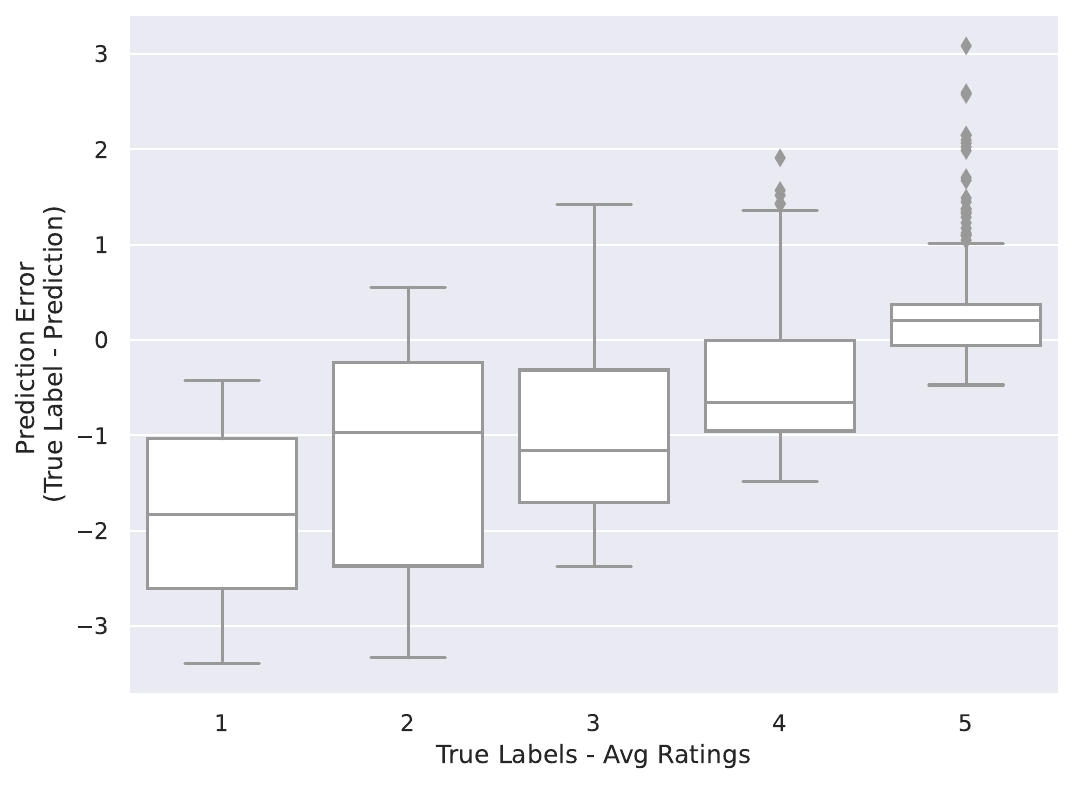}
        \caption{XLM-R OSN Errors}
        \label{fig:errors-prediction-osn}
    \end{subfigure}

    \begin{subfigure}{0.32\textwidth}
        \includegraphics[width=\textwidth]{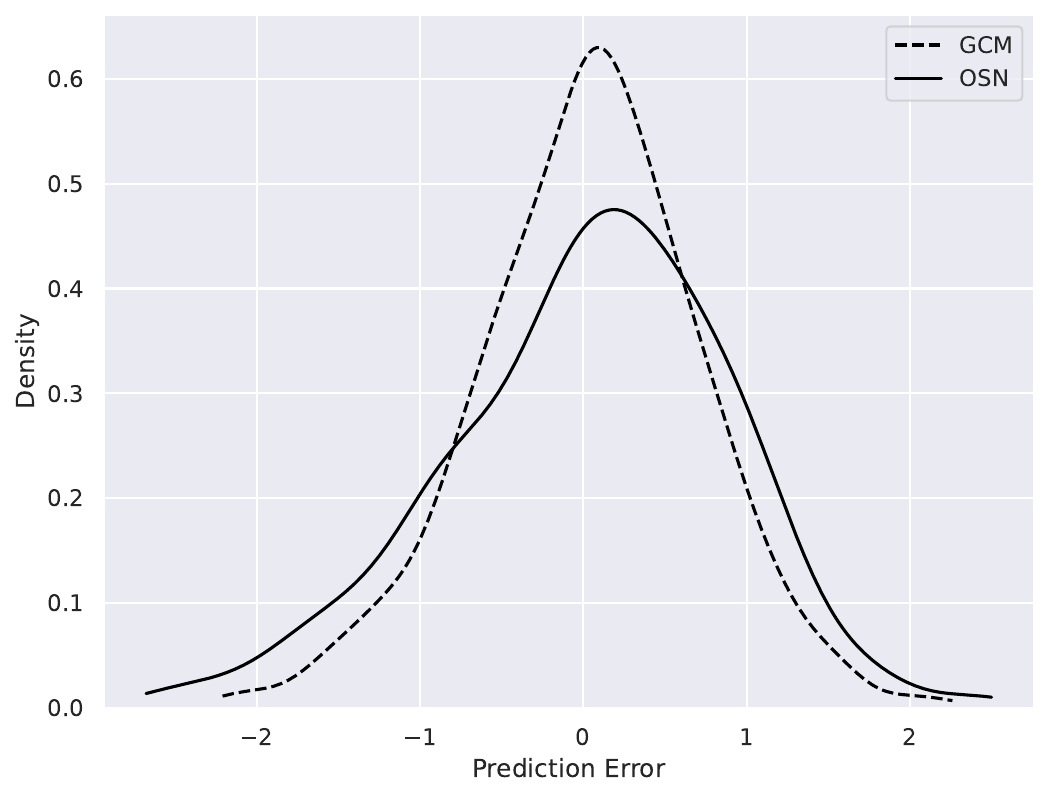}
        \caption{LLama 3.2-3B Distribution of Errors}
        \label{fig:error-dist-gcm-osn-llama}
    \end{subfigure}
    \begin{subfigure}{0.32\textwidth}
        \includegraphics[width=\textwidth]{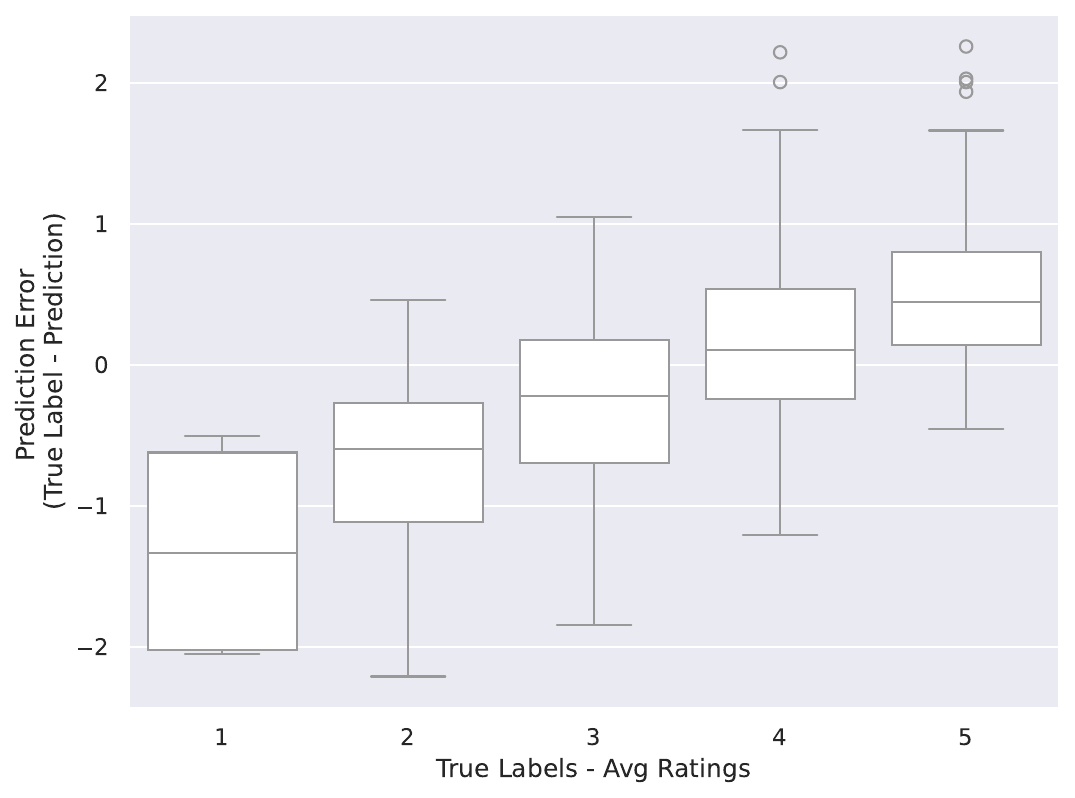}
        \caption{LLama 3.2-3B GCM Errors}
        \label{fig:errors-prediction-gcm-llama}
    \end{subfigure}
    \begin{subfigure}{0.32\textwidth}
        \includegraphics[width=\textwidth]{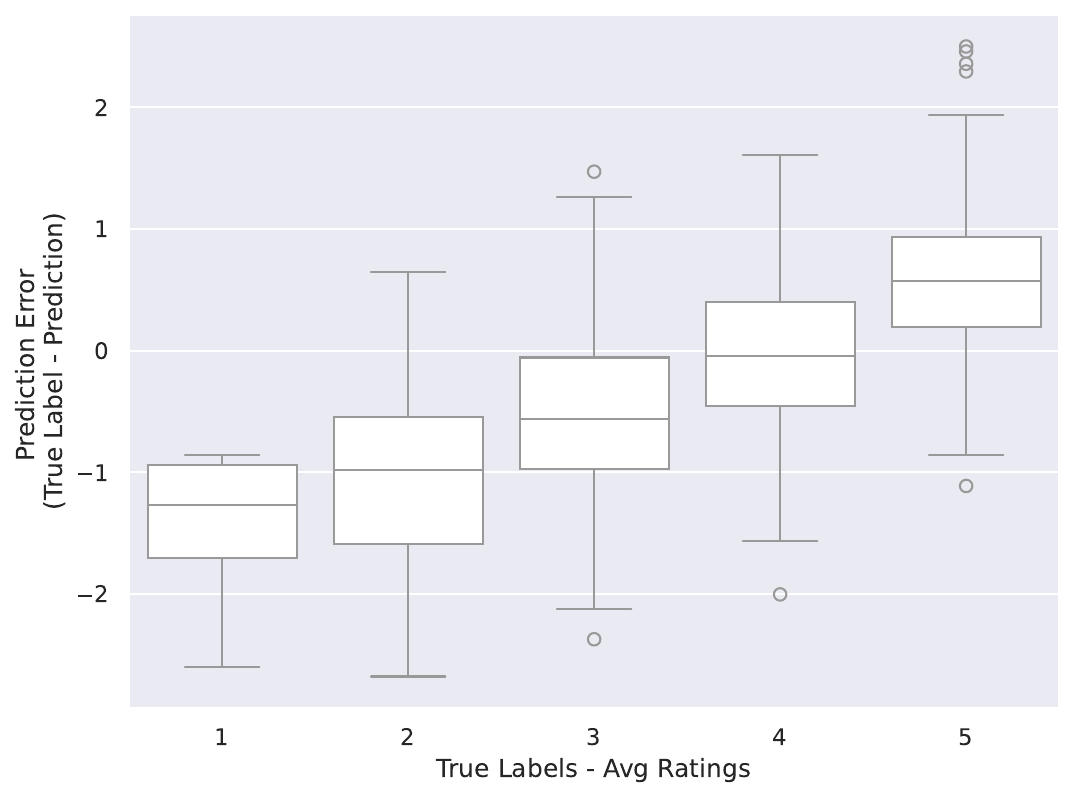}
        \caption{LLama 3.2-3B OSN Errors}
        \label{fig:errors-prediction-osn-llama}
    \end{subfigure}

\label{fig:errors-dist}
    \caption{(a) \& (d) Distribution of errors for GCM and OSN. Peak around zero - for both OSN and GCM - indicate model's performance. Long tail and skew towards left indicate propensity of model to over-predict; (b) \& (c) show distribution of prediction errors across Average Ratings labels - GCM and OSN respectively for XLM-R, while (e) \& (f) show the distribution for fine-tuned Llama 3.2-3B. Both models over-predicts for less acceptable sentences, and predicts acceptable sentences more correctly.}

\end{figure*}

To analyze the prediction errors made by the model, we examine the errors for the two best-performing model—fine-tuned XLM-R and Llama 3.2 — on both the OSN and GCM datasets. Prediction error is calculated as the difference between the true label and the model's prediction, with a negative error indicating that the model's prediction exceeds the average human rating. \hbox{Figure~\ref{fig:error-dist-gcm-osn}, \ref{fig:error-dist-gcm-osn-llama}} illustrates the distribution of prediction errors for the OSN and GCM test sets for fine-tuned XLM-R and Llama 3.2-3B models. Both models exhibit a peak around zero, indicating a reasonable prediction accuracy. However, XLM-R shows a more pronounced left skew, suggesting a higher tendency to over-predict compared to LLaMA 3.2-3B. Majority of errors fall within the range of \hbox{$[-1, 1]$} for both datasets. However, OSN predictions show a higher proportion of errors exceeding \hbox{$\pm1$} compared to GCM predictions. Figures\hbox{~\ref{fig:errors-prediction-gcm}, \ref{fig:errors-prediction-gcm-llama}} and\hbox{~\ref{fig:errors-prediction-osn}, \ref{fig:error-dist-gcm-osn-llama}} depict the distribution of errors across different scales of human average ratings for the GCM and OSN datasets, respectively. Both models tend to over-predict for lower rating categories (less acceptable sentences, i.e., sentences with an average human rating below 3), but LLaMA 3.2-3B exhibits a more balanced error distribution across ratings, with fewer extreme outliers compared to XLM-R. Furthermore, LLaMA 3.2-3B appears to have a slightly lower variance in errors, particularly for higher ratings, suggesting that it may be more reliable in predicting highly acceptable sentences. Overall, while both models demonstrate over-prediction tendencies for lower-rated samples, LLaMA 3.2-3B appears to provide more stable predictions with reduced skew and variance.

To further analyse the errors made by models, we explore the relationship between the error magniture and various code-mixing metrics. Table\hbox{~\ref{tab:error_analysis_correlation}} shows that none of the code-mixing metrics exhibit strong correlations with prediction error for either dataset, though OSN has slightly higher correlation values (e.g., CMI: 0.012 vs. -0.0035, SyMCoM Sentence Scores: 0.049 vs. -0.033), suggesting a marginal relationship between these metrics and model performance in OSN. In Table\hbox{~\ref{tab:error_analysis_correlation}}, we explore potential monotonic relationships between prediction errors and code-mixing metrics. The SyMCoM score is computed across all Parts of Speech (PoS) tags, although the way a specific PoS tag is mixed might still correlate with instances that challenge the model. In Table\hbox{~\ref{tab:error_analysis_symcom_anova}}, the ANOVA analysis shows that certain PoS categories significantly influence SyMCoM scores, but the affected categories differ between datasets. To perform the One-way ANOVA analysis, we transform the PoS tag-specific SyMCoM scores into a categorical variable. This variable takes on the values of Monolingual, Mixed, or Absent, signifying whether tokens for a PoS tag originate exclusively from one language, from both languages or are not present within the sentence. GCM exhibits significant effects for ADP (F = 3.055, p = 0.048), PPRON (F = 3.584, p = 0.078), and VERB (F = 2.449, p = 0.087), whereas OSN shows significance for NOUN (F = 3.832, p = 0.022), indicating that nouns play a stronger role in code-mixing for OSN while pronouns and verbs are more influential in GCM. This suggests that linguistic structures influence code-mixing differently across datasets, which may impact model performance and generalizability.

\input{tables/error_analysis}

\subsection{Zero-shot transfer across languages}

\input{tables/07_ente_transfer}

The findings in Table\hbox{~\ref{tab:en-te-transfer}} highlight the effectiveness of transferring acceptability models trained on en-hi (English-Hindi) data to the en-te (English-Telugu) language pair. The random baseline performs the worst, with an RMSE of 1.76 and MAE of 1.44. The human baseline achieves the best performance, with an RMSE of 0.74 and MAE of 0.55, serving as an upper bound for model performance. All models in the zero-shot setting (XLM-R, mBART, and Llama 3.2 - 3B) perform poorly, with high RMSE and MAE values, indicating that without prior training on related data, the models struggle to generalize to the en-te language pair. Models fine-tuned on en-hi data significantly outperform their zero-shot counterparts, demonstrating effective transferability across related code-mixed language pairs. Among the transferred models, Llama 3.2 - 3B performs the best, achieving an RMSE of 1.09 and MAE of 1.01. XLM-R and mBART also show considerable improvement, with RMSE values of 1.28 and 1.29, respectively, and MAE values of 1.06 and 1.02. Overall, fine-tuning on en-hi data leads to significant performance improvements for en-te, underscoring the cross-lingual transferability of models trained on related language pairs.
Llama 3.2 - 3B demonstrates the strongest transfer performance, suggesting its robustness and capacity for cross-lingual generalization in code-mixed settings.

%% file: tables/03-results_new.tex
\begin{table}
\centering
\caption{Performance measure for GCM and OSN data, using different baselines and models. Fine-tuned MLLMs outperform human baselines, and regression models trained on only code-mix metrics. $\downarrow$ indicates lower number shows better performance.}
\label{tab:results_seperate_training_withChatGPT}
\begin{tabular}{llcccc} 
\hline\hline
                                                                                      &                                & \multicolumn{2}{c}{GCM}                                                      & \multicolumn{2}{c}{OSN}                                                       \\ 
\hline
                                                                                      &                                & \multicolumn{1}{l}{RMSE $\downarrow$} & \multicolumn{1}{l}{MAE $\downarrow$} & \multicolumn{1}{l}{RMSE $\downarrow$} & \multicolumn{1}{l}{MAE $\downarrow$}  \\ 
\hline\hline
Random Predictions                                                                    &                                & 1.54                                  & 1.25                                 & 1.78                                  & 1.47                                  \\
Human Annotations Baseline                                                            &                                & 1.05                                  & 0.78                                 & 0.95                                  & 0.78                                  \\ 
\hline
\multicolumn{6}{c}{Predictors Trained using Code-Mixing Metrics}                                                                                                                                                                                                                      \\ 
\hline
\begin{tabular}[c]{@{}l@{}}Feed Forward Network\\Using~ Code-Mix Metrics\end{tabular} &                                & 0.97                                  & 0.83                                 & 1.42                                  & 1.01                                  \\
~~~~~~~~~~ + MLLM Perplexity Scores                                                   &                                & 0.97                                  & 0.84                                 & 1.52                                  & 1.04                                  \\ 
\hline
\multicolumn{6}{c}{Fine-Tuning Pre-Trained Models}                                                                                                                                                                                                                                    \\ 
\hline
\multirow{2}{*}{XLM-Roberta}                                                          & Romanised                      & 0.8                                   & 0.62                                 & 1.08                                  & 0.78                                  \\
                                                                                      & Normalised                     & 0.74                                  & 0.58                                 & 1.01                                  & 0.743                                 \\
\multirow{2}{*}{BERNICE}                                                              & Romanised                      & 0.79                                  & 0.62                                 & 1.02                                  & 0.77                                  \\
                                                                                      & Normalised                     & 0.72                                  & 0.57                                 & 1.05                                  & 0.78                                  \\
\multirow{2}{*}{IndicBERT}                                                            & Romanised                      & 0.88                                  & 0.7                                  & 1.17                                  & 0.94                                  \\
                                                                                      & Normalised                     & 0.87                                  & 0.69                                 & 1.2                                   & 0.97                                  \\ 
\hline
\multirow{2}{*}{mBART}                                                                & Romanised                      & 0.84                                  & 0.70                                 & 1.04                                  & 0.85                                  \\
                                                                                      & Normalised                     & 0.70                                  & 0.54                                 & 0.94                                  & 0.70                                  \\
\multirow{2}{*}{mT5}                                                                  & Romanised                      & 0.95                                  & 0.81                                 & 1.11                                  & 0.94                                  \\
                                                                                      & Normalised                     & 0.93                                  & 0.80                                 & 1.16                                  & 0.93                                  \\ 
\hline
\multirow{2}{*}{Llama 3.2 - 1B}                                                       & Romanised                      & 0.851                                 & 0.69                                 & 0.96                                  & 0.76                                  \\
                                                                                      & Normalised                     & 0.70                                  & 0.55                                 & 0.90                                  & 0.70                                  \\
\multirow{2}{*}{Llama 3.2 - 3B}                                                       & Romanised                      & 0.86                                  & 0.70                                 & 0.98                                  & 0.76                                  \\
                                                                                      & Normalised                     & \textbf{0.68}                         & \textbf{0.53}                        & \textbf{0.87}                         & \textbf{0.69}                         \\
\multirow{2}{*}{Qwen 2.5 1..5B}                                                       & Romanised                      & 0.89                                  & 0.72                                 & 1.01                                  & 0.80                                  \\
                                                                                      & Normalised                     & 0.75                                  & 0.59                                 & 1.04                                  & 0.81                                  \\
\multirow{2}{*}{Qwen 2.5 3B}                                                          & Romanised                      & 0.89                                       & 0.71                                     & 1.01                                      & 0.79                                       \\
                                                                                      & Normalised                     & 0.79                                  & 0.63                                 & 0.99                                      & 0.77                                       \\
\multirow{2}{*}{Phi 3 - 3B}                                                           & Romanised                      & 0.90                                  & 0.73                                 & 1.05                                  & 0.83                                  \\
                                                                                      & Normalised                     & 0.74                                  & 0.58                                 & 0.99                                  & 0.79                                  \\ 
\hline
\multicolumn{6}{c}{Zero/Few-shot Prompting LLMs}                                                                                                                                                                                                                                      \\ 
\hline
\multirow{2}{*}{ChatGPT-3.5 - ZeroShot}                                               & \multicolumn{1}{c}{Romanised}  & 1.20                                  & 0.95                                 & 1.28                                  & 1.02                                  \\
                                                                                      & \multicolumn{1}{c}{Normalised} & 1.22                                  & 0.96                                 & 1.39                                  & 1.12                                  \\
\multirow{2}{*}{ChatGPT-3.5 - 5-Shot}                                                 & \multicolumn{1}{c}{Romanised}  & 1.16                                  & 0.94                                 & 1.75                                  & 1.48                                  \\
                                                                                      & \multicolumn{1}{c}{Normalised} & 1.09                                  & 0.87                                 & 1.71                                  & 1.46                                  \\
\hline\hline
\end{tabular}
\end{table}

%% file: tables/tokenfertility_perplexity.tex
\begin{table}[!b]
\centering
\caption{Tokenizer Fertility and Perplexity/Surprisal comparision for normalized and romanised code-mixed samples. Perplexity/Surprisal values clearly show that the models are less surprised for normalised code-mixed sentences compared to romanised. Tokenizer fertility shows that for Encoder and Encoder-Decoder models normalised sentences are broken into lesser tokens compared to romanised, while Decoder models shows the opposite trend.}
\label{tab:tokenizerfertility_perplexity}
\begin{tabular}{lllllllll} 
\hline\hline
\multicolumn{1}{c}{\multirow{3}{*}{\textbf{Model}}} & \multicolumn{4}{c}{\textbf{Tokenizer Fertility }}                                             & \multicolumn{4}{c}{\textbf{Perplexity/Surprisal }}                    \\ 
\cline{2-9}
\multicolumn{1}{c}{}                                & \multicolumn{2}{c}{\textbf{GCM }}             & \multicolumn{2}{c}{\textbf{OSN }}             & \multicolumn{2}{c}{\textbf{GCM }} & \multicolumn{2}{c}{\textbf{OSN}}  \\ 
\cline{2-9}
\multicolumn{1}{c}{}                                & \textbf{Norm}         & \textbf{Roman}        & \textbf{Norm}         & \textbf{Roman}        & \textbf{Norm} & \textbf{Roman}    & \textbf{Norm} & \textbf{Roman}    \\ 
\hline\hline
BERNICE                                               & 1.57                  & 1.61                  & 1.86                  & 1.97                  & 39.2          & 70.5              & 148.0         & 189.5             \\
XLM-Roberta                                               & 1.53                  & 1.73                  & 1.80                  & 1.97                  & 44.1          & 69.7              & 122.5         & 133.4             \\
IndicBERT                                           & 1.56                  & 1.81                  & 1.82                  & 1.99                  & 77.2          & 99.2              & 324.6         & 429.3             \\
mT5                                                 & 1.47                  & 1.64                  & 1.82                  & 2.0                   & 730.2         & 2905.9            & 1818.1        & 8480.2            \\
mBART                                               & 1.53                  & 1.73                  & 1.80                  & 1.97                  & 4212.5        & 8940.8            & 1006.1        & 2538.8            \\
Llama 3.2 1B                                        & \multirow{2}{*}{1.76} & \multirow{2}{*}{1.57} & \multirow{2}{*}{2.19} & \multirow{2}{*}{2.0}  & 211.8         & 1870.4            & 250.2         & 555.0             \\
Llama 3.2 3B                                        &                       &                       &                       &                       & 192.7         & 1613.5            & 222.9         & 443.4             \\
Qwen 2.5 1.5B                                       & \multirow{2}{*}{2.43} & \multirow{2}{*}{1.48} & \multirow{2}{*}{2.95} & \multirow{2}{*}{1.98} & 53.8          & 2478.9            & 76.8          & 713.3             \\
Qwen 2.5 3B                                         &                       &                       &                       &                       & 69.2          & 1698.2            & 82.3          & 965.3             \\
Phi 3 3B                                            & 2.8                   & 1.6                   & 3.5                   & 2.3                   & 33.1          & 2139.2            & 46.2          & 516.8             \\
\hline\hline
\end{tabular}
\end{table}

%% file: tables/error_analysis.tex
\begin{table}[t]
    \begin{minipage}[t]{0.48\textwidth}

        \raggedright
        \caption{Correlation between fine-tuned Llama 3.2-3B prediction error and code-mix metrics indicate that there aren't any clear correlation between the prediction error and the code-mix metrics of the samples.}
        \label{tab:error_analysis_correlation}

        \resizebox{\linewidth}{!}{%
            \begin{tabular}{lrr} 
            \hline\hline
                                   & \multicolumn{1}{l}{GCM} & \multicolumn{1}{l}{OSN}  \\ 
            \hline\hline
            Length                 & -0.039                  & 0                        \\
            CMI                    & -0.0035                 & 0.012                    \\
            SPAvg                  & 0.054                   & 0.038                    \\
            Burstiness             & -0.019                  & 0                        \\
            SyMCoM Sentence Scores & -0.033                  & 0.049                    \\
            Surprisal              & 0.08                    & 0.2                      \\
            \hline\hline
            \end{tabular}
    
        }

    \end{minipage}%
    \quad
    \begin{minipage}[t]{0.45\textwidth}
        \raggedleft
        \caption{One way ANOVA analysis between PoS tag wise SyMCoM scores and prediction error of fine-tuned Llama 3.2-3B. We convert SyMCoM values into categories to encode if a particular PoS tag is mixed or monolingual or not present in the text.}
        \label{tab:error_analysis_symcom_anova}            
        \resizebox{\linewidth}{!}{%

\begin{tabular}{lllll} 
\hline\hline
      & \multicolumn{2}{c}{\textbf{GCM}}                                                      & \multicolumn{2}{c}{\textbf{OSN}}                                                       \\ 
\hline\hline
      & F-statistic                               & p-value                                   & F-statistic                               & p-value                                    \\
ADJ   & {\cellcolor[rgb]{0.984,0.714,0.725}}1.794 & {\cellcolor[rgb]{0.973,0.506,0.518}}0.167 & {\cellcolor[rgb]{0.984,0.796,0.808}}1.325 & {\cellcolor[rgb]{0.976,0.561,0.569}}0.267  \\
ADP   & {\cellcolor[rgb]{0.976,0.502,0.51}}3.055* & {\cellcolor[rgb]{0.973,0.424,0.431}}0.048 & {\cellcolor[rgb]{0.984,0.82,0.831}}1.161  & {\cellcolor[rgb]{0.976,0.592,0.6}}0.314    \\
ADV   & {\cellcolor[rgb]{0.988,0.969,0.98}}0.273  & {\cellcolor[rgb]{0.984,0.925,0.937}}0.761 & {\cellcolor[rgb]{0.988,0.847,0.859}}0.989 & {\cellcolor[rgb]{0.976,0.627,0.635}}0.373  \\
AUX   & {\cellcolor[rgb]{0.988,0.969,0.98}}0.270  & {\cellcolor[rgb]{0.984,0.925,0.937}}0.764 & {\cellcolor[rgb]{0.988,0.922,0.933}}0.497 & {\cellcolor[rgb]{0.98,0.773,0.784}}0.609   \\
CCONJ & {\cellcolor[rgb]{0.988,0.988,1}}0.147     & {\cellcolor[rgb]{0.984,0.882,0.894}}0.702 & {\cellcolor[rgb]{0.984,0.835,0.847}}1.057 & {\cellcolor[rgb]{0.976,0.612,0.62}}0.348   \\
DET   & {\cellcolor[rgb]{0.988,0.91,0.922}}0.632  & {\cellcolor[rgb]{0.98,0.765,0.773}}0.531  & {\cellcolor[rgb]{0.98,0.627,0.635}}2.433  & {\cellcolor[rgb]{0.973,0.451,0.459}}0.089  \\
INTJ  & {\cellcolor[rgb]{0.98,0.631,0.643}}2.283  & {\cellcolor[rgb]{0.973,0.482,0.49}}0.131  & {\cellcolor[rgb]{0.988,0.988,1}}0.050     & {\cellcolor[rgb]{0.988,0.988,1}}0.951      \\
NOUN  & {\cellcolor[rgb]{0.984,0.82,0.831}}1.160  & {\cellcolor[rgb]{0.976,0.612,0.62}}0.314  & {\cellcolor[rgb]{0.973,0.412,0.42}}3.832* & {\cellcolor[rgb]{0.973,0.412,0.42}}0.022   \\
NUM   & {\cellcolor[rgb]{0.984,0.804,0.816}}1.256 & {\cellcolor[rgb]{0.976,0.592,0.6}}0.285   & {\cellcolor[rgb]{0.984,0.816,0.827}}1.199 & {\cellcolor[rgb]{0.976,0.565,0.576}}0.274  \\
PART  & {\cellcolor[rgb]{0.988,0.969,0.98}}0.269  & {\cellcolor[rgb]{0.984,0.925,0.937}}0.765 & {\cellcolor[rgb]{0.988,0.871,0.882}}0.833 & {\cellcolor[rgb]{0.976,0.667,0.675}}0.435  \\
PRON  & {\cellcolor[rgb]{0.988,0.945,0.957}}0.423 & {\cellcolor[rgb]{0.984,0.851,0.863}}0.655 & {\cellcolor[rgb]{0.98,0.616,0.624}}2.506  & {\cellcolor[rgb]{0.973,0.447,0.455}}0.082  \\
PPRON & {\cellcolor[rgb]{0.973,0.412,0.42}}3.584* & {\cellcolor[rgb]{0.973,0.412,0.42}}0.028  & {\cellcolor[rgb]{0.988,0.875,0.886}}0.809 & {\cellcolor[rgb]{0.976,0.675,0.682}}0.446  \\
SCONJ & {\cellcolor[rgb]{0.988,0.988,1}}0.164     & {\cellcolor[rgb]{0.988,0.988,1}}0.849     & {\cellcolor[rgb]{0.98,0.643,0.655}}2.315  & {\cellcolor[rgb]{0.973,0.459,0.467}}0.100  \\
VERB  & {\cellcolor[rgb]{0.98,0.604,0.612}}2.449* & {\cellcolor[rgb]{0.973,0.451,0.459}}0.087 & {\cellcolor[rgb]{0.988,0.953,0.965}}0.287 & {\cellcolor[rgb]{0.984,0.863,0.875}}0.751  \\
\hline\hline
\end{tabular}

        }

    \end{minipage}%
\end{table}

%% file: tables/07_ente_transfer.tex
\begin{table}[!t]
\centering
\caption{Transfer of en-hi acceptability model to en-te data. Results show that model trained on en-hi data is better than random baselines, and zero-shot XLM-R, mBART, LLama 3.2-3B, indicating tranfer from en-hi language pair to en-te. \hbox{$\downarrow$} indicates lower number shows better performance.}
\label{tab:en-te-transfer}
\begin{tabular}{lll} 
\hline\hline
                                                                                      & RMSE $\downarrow$ & MAE $\downarrow$  \\ 
\hline\hline
Random Baseline                                                                       & 1.76              & 1.44              \\
Human Baselines                                                                       & 0.74              & 0.55              \\ 
\hline
XLM-R - Zero-Shot                                                                     & 3.9               & 3.7               \\
\begin{tabular}[c]{@{}l@{}}XLM-R - Transfer from \\Training on en-hi\end{tabular}     & 1.28              & 1.06              \\ 
\hline
mBART Zero-Shot                                                                       & 3.5               & 3.2               \\
\begin{tabular}[c]{@{}l@{}}mBART Transfer from\\Training on en-hi\end{tabular}        & 1.29              & 1.02              \\ 
\hline
Llama 3.2 3B Zero-Shot                                                                & 3.2               & 3.0               \\
\begin{tabular}[c]{@{}l@{}}Llama 3.2 3B Transfer from\\Training on en-hi\end{tabular} & 1.09              & 1.01              \\
                                                                                      &                   &                   \\
\hline\hline
\end{tabular}
\end{table}



%% file: sections/06-conclusion.tex
\section{Discussion and Conclusion}\label{sec:06_conclusion}
In this work, we propose a novel Hindi-English code-mixed dataset, Cline, annotated with human judgements for acceptability, capturing a crucial aspect of bilingualism. Acceptability judgements enable quality assessment of code-mixed text, unlocking data augmentation and improved generation, analysis capabilities. We demonstrate that conventional code-mixing complexity metrics, commonly employed for curating code-mixed resources, exhibit limited correlation with human acceptability judgements of code-mixed text. This underscores the necessity for improved methods in evaluating the acceptability of code-mixed text. Our analysis shows evidence that pre-trained multilingual LLMs like XLM-R and Bernice encode some linguistic information for acceptability of complex cross-lingual phenomena such as code-mixing. By regressing for average acceptability rating against metrics and MLLM's perplexity (we observed no significant relationship), we gather indirect evidence towards the complexity involved in computationally modeling code-mixed acceptability. Fine-tuning encoder-only MLLMs, such as Llama3.2, for predicting acceptability ratings on our dataset outperforms the zero- and few-shot capability of much larger instruction-following LLMs like ChatGPT. Acceptability predictors trained using Cline~ can be used to curate larger quality-controlled code-mixed corpora, enabling data-intensive approaches for computational research in code-mixed settings. In code-mixed text generation pipelines, the ability to assess the acceptability of generated sentences can significantly enhance the overall quality of code-mixed text generation.

\textbf{Broader Applications and Implications}: Our work is particularly relevant to NLP applications, such as interactive agents and language learning/teaching, especially in multilingual societies like India, where the ability to generate and process code-mixed text is crucial. Understanding and processing code-mixed content on platforms like Twitter, Facebook, and Instagram to analyze sentiment, detect trends, gather user insights, and detect harmful content, hate speech, and policy violations in code-mixed text on social platforms, which might otherwise evade monolingual content filters. Speakers might have differing levels of bilingual proficiency, and in such cases code-mixed text/speech can act as the bridge.~\hbox{\citet{Bawa2020DoMU}} demonstrate that multilingual users prefer chatbots that code-mix and code-mix like the users. Enabling customer-facing agents - voice assistants and chatbots- to understand and respond appropriately to users who naturally switch between languages during conversation can improve customer experience. Code-mixing in teaching can benefit students by making learning more accessible, engaging, and culturally relevant by allowing teachers to utilize familiar language elements from students' native tongue to explain complex concepts, bridge language gaps, and create a more inclusive classroom environment, especially when dealing with diverse linguistic backgrounds. For scaling all the aformentioned applications, language models are central to generating and analyzing code-mixed text. Given their resource-scarce nature, quality-controlled synthetic data is highly beneficial. Acceptability of code-mixed sentences, as presented in this study, can help in curating high-quality code-mixed corpus, which can be used for adapting a language model to code-mixed settings. Such model adaptation benefits both generation and analysis capabilities for code-mixed text.

\subsection{Limitations \& Future Work}
Firstly, the current study is situated only in the context of English-Hindi code-mixed texts, the patterns observed in one language pair may not hold for other language pairs such as English-French or French-Arab. A multilingual investigation into acceptability of code-mixing is thus imperative. However, the lack of high-quality corpora, LLMs and task-specific models, viz., Language Identifier and PoS Tagger in extremely low-resource settings, poses a major challenge for extensions of the current work. Our work defines the problem of acceptability in code-mixed text and proposes a framework for creating data resources to address this task, which we hope will serve as a foundation for the broader research community to build upon. Developing a generalized model of code-mixed text acceptability across diverse language pairs is a crucial long-term goal for this research direction. As noted, patterns of code-mixing vary significantly across language pairs—e.g., the switching patterns in English-Spanish code-mixed text differ greatly from those in English-Hindi. Moreover, code-mixing is a complex sociolinguistic phenomenon influenced by numerous factors beyond the surface-level form of a given utterance (e.g., syntactic, semantic, and sociolinguistic considerations)~\cite{Bullock_Toribio_2009, auer1984bilingual, milroy2008sociolinguistics}. These intricacies make creating a generic, universally applicable model of code-mixed text acceptability an ambitious but challenging endeavor. In this work, we have introduced datasets for two language pairs (English-Hindi and English-Telugu) and explored whether there is evidence of transferability between these pairs. However, we do not claim that the datasets or models trained on one language pair are universally applicable across all language pairs. Instead, we see this work as an important first step in addressing this complex problem and anticipate that future research will build upon these initial contributions to create more universal resources and tools for analyzing code-mixed text acceptability. We believe that contributing resources for two low-resource, data-constrained language pairs is a significant and meaningful step given the nature of the problem.

Secondly, it has been proven that domain adaptation improves the performance of LLMs on downstream tasks. However, the current analysis focuses solely on pre-trained LLMs; hence, the impact of domain adaptation (continued pretraining) on acceptability performance in a code-mixed corpus is not examined.

Finally, the crowd-sourced annotations are conducted on a university campus with undergraduate students, resulting in a skewed demographic for the acceptability ratings. While the linguistic backgrounds of the annotators were controlled, the current ratings might still be bounded by the skew in rater demographics. We acknowledge that acceptability judgments can indeed vary based on factors such as age, region, and language proficiency, and such judgments may vary across demographics. This work does
not profess to capture the full complexity of the code-mixed text acceptability phenomenon.
It is inherently limited by annotator pool, which is a known challenge in any annotation task.
However, the primary goal of this work is to define the problem space and establish a foundational framework for analyzing the acceptability of code-mixed text. While this study represents an important step forward, we believe a more unifying theory or model of code-mixed text acceptability will emerge as the research community builds upon this work, incorporating diverse perspectives and additional methodologies. Future work could examine how age, bilingual proficiency, and regional variations shape these judgments by expanding the annotator pool to include a diverse range of participants. For instance, a controlled study with stratified sampling could assess whether younger individuals or those schooled in bilingual environments show higher acceptance of code-mixing compared to older individuals with formal monolingual education. Similarly, a self-reported bilingual proficiency test could help analyze whether active bilinguals have more lenient acceptability thresholds than passive bilinguals.

Additionally, regional factors may play a role in shaping preferences, particularly in urban vs. rural settings. Individuals in urban areas, where multilingual interactions are more common, might exhibit greater inclination for code-mixing than those in linguistically homogeneous rural communities. This could be explored through a geographically distributed annotation study, tracking how exposure to a dominant language in a given region correlates with acceptability judgments. Finally, the linguistic structure of a language itself may influence perceptions—languages with a history of borrowing and flexible grammar structures might foster higher code-mixing acceptance than those with strict linguistic norms. A comparative cross-linguistic study could help identify such trends, shedding light on the deeper cognitive and social factors driving code-mixing acceptability.

In our work we have incorporated syntactic features like PoS tags combined with language IDs. Furthermore, when we utilize pretrained multilingual language models, we expect the models to capture syntactic and semantic nuances of the code-mixed samples. Code mixing is a sociolinguistic phenomenon and is affected by multiple parameters that are beyond the form manifestation of a particular code-mixed utterance. Code-Mixing embodies, or corresponds with, a wide range of sociolinguistic factors that interact or operate simultaneously. Given the multitude of variations within and across language pairs, a generic model of acceptability of code-mixed text is an extremely worthwhile but at the same time equally hard problem. Incorporating contextual insights – demographic variables of who is speaking, pragmatic context of the code-mixed utterance is an excellent suggestion, and we leave this to future work. We have added this discussion as part of future work and limitations sections in the revised manuscript.

We foresee the usefulness of datasets like Cline~ in enhancing downstream Natural Language Generation tasks. This enhancement can lead to more natural communication between humans and machines by alleviating monolingual constraints imposed on users. Future work might explore unsupervised approaches to code-mixed acceptability quantification and investigate the scope of transfer learning in adapting LLMs' monolingual acceptability knowledge to complex settings like code-mixed generation. This work also has implications for generating and curating high-quality filtered datasets based on the acceptability predictors trained on crowd-sourced judgements. Acceptability predictors can be directly plugged into data curation pipelines for code-mixed text to weed out undesired and unnatural samples. Essentially this can further improve the domain adaptation of LLMs on code-mixed text by aligning LLMs towards language deemed more acceptable by bilinguals. Additionally, filtering and improving code-mix benchmarks like GLUECoS \cite{khanuja-etal-2020-gluecos} and LinCE \cite{aguilar-etal-2020-lince} to investigate performance improvements in downstream tasks is an exciting avenue for future work. Further exploration of zero-shot transfer to other code-mixed language pairs can be fruitful.

Recent works like \citet{fornaciari2021beyond, 10.1613/jair.1.12752} have conjectured that annotator disagreement can aid model training. Predicting the probability distribution over annotator labels in addition to the average rating can act as a regularizer and prevent overfitting. A similar paradigm can help boost performance on ambiguous annotations in subjective tasks like acceptability judgements.

Finally, we believe the crowd-sourced annotations, along with the acceptability predictors, can assist in the creation of probing benchmarks for code-mixed knowledge in pre-trained multilingual LLMs.

Exploring interpretability and understanding the factors that make a sentence acceptable or unacceptable remains an important direction for future research. Techniques such as LIME, SHAP, and attention visualization, while widely used, face significant limitations when applied to transformer-based models. These include shallow or unreliable explanations, oversimplified surrogate models, computational challenges, and the lack of a direct correlation between attention weights and feature importance. Additionally, the current dataset provides binary acceptability judgments without detailed rationales or features, limiting the ability to identify precise linguistic or contextual drivers of model decisions. Future work could focus on developing datasets with richer annotations that capture the reasoning behind acceptability judgments and creating advanced explainability methods tailored to transformer architectures.

%% file: sections/appendix.tex
\section{Code-Mixing Metrics}\label{appendix:codemix_metrics}

One way to quantify code-mixing is to treat a code-mixed utterance just as a series of language identifiers, ignoring other syntactic, semantic properties. Code Mixing Index (CMI), Number of Switch Points, Burstiness are metrics computed based on just the token-wise language ID tags. CMI~\hbox{\citep{gamback-das-2016-comparing}} is a ratio-based metric (ratio of tokens from Language 1 and Language 2). CMI is defined in Equation\hbox{~\ref{CMI}}, where $n$ is total number of tokens and $u$ is the number of language independent tokens, $n-u$ is the sum of number of tokens from N languages and $max\{w_i\}$ is the highest number of words belonging to a particular language.
\begin{equation} \label{CMI}
  CMI =
    \begin{cases}
      100 \times \left [ 1 - \frac{max\left \{ w_i \right \}}{n-u} \right ] & \text{if $n > u$}\\
      0 & \text{if $n = u$}
    \end{cases}       
\end{equation}

Number of Switch Points~(SP) captures the number of times the language was switched in an utterance. \hbox{~\citet{guzman17_interspeech}} presented multiple metrics which also take into account the time ordering of language IDs - Burstiness being one of them, defined in Equation\hbox{~\ref{burstiness}} where $\sigma_\tau$ denote the standard deviation of the language spans and $m_\tau$ the mean of the language spans. Burstiness is bounded within the interval [-1, 1]. Corpora with anti-bursty, periodic dispersions of switch points take on Burstiness values closer to -1. By contrast, corpora with less predictable patterns of switching take on values closer to 1.
\begin{equation} \label{burstiness}
Burstiness = \frac{\left ( \sigma_\tau - m_\tau  \right )}{\left ( \sigma_\tau + m_\tau \right )}
\end{equation}

Metrics of Code-Mixing which rely solely on token wise language IDs do not encode the distribution of syntactic units among the languages mixed in a sentence i.e which language contributed which Parts of Speech tags in a code-mixed sentence. SyMCoM\hbox{~\citep{kodali-etal-2022-symcom}} quantifies which language is contributing which syntactic unit~(PoS tags for instance), providing means to encode the degree of mixing within and across syntactic units in a sentence. SyMCoM for each PoS tag is defined as in Equation\hbox{~\ref{SyMCoM}}, where SU~(Syntactic Unit) is one of the PoS tag, L1 and L2 are the languages that are mixed. SyMCoM takes into account both language ID and syntactic category of a token. SyMCoM of each PoS tag can be further aggregated to have a single SyMCoM score for a given sentence.
\begin{equation} \label{SyMCoM}
SyMCoM_{SU} = \frac{(Count_{SU_{L1}}) - (Count_{SU_{L2}})}{\sum_{i = 1}^{2} {Count_{SU_{Li}}}}
\end{equation}